\documentclass{article}

\usepackage[preprint]{neurips_2021}

\usepackage[american]{babel}
\usepackage[utf8]{inputenc} 
\usepackage[T1]{fontenc}    
\usepackage{hyperref}       
\usepackage{url}            
\usepackage{booktabs}       
\usepackage{amsfonts}       
\usepackage{nicefrac}       
\usepackage{microtype}      
\usepackage{xcolor}         

\usepackage{natbib} 

\usepackage{mathtools} 
\usepackage{tikz} 

\usepackage{microtype}
\usepackage{graphicx}
\usepackage{subfigure}
\usepackage{booktabs} 

\usepackage{hyperref}
\usepackage{algorithm2e}

\usepackage{hyperref}       
\usepackage{url}            
\usepackage{amsfonts}       
\usepackage{nicefrac}       
\usepackage{microtype}      
\usepackage{amsmath}
\usepackage{amssymb}
\usepackage{wrapfig}
\usepackage{amsthm}
\usepackage{mathtools}
\usepackage{bm}
\usepackage{dsfont}
\usepackage{mathrsfs}

\usepackage[capitalize]{cleveref}
\usepackage{pgfplots}

\newtheorem{remark}{Remark}

\newcommand{\E}{\ensuremath{\mathbb{E}}}
\newcommand{\PP}{\ensuremath{\mathbb{P}}}

\newcommand{\hp}{\ensuremath{\widehat{p}}}

\newcommand{\field}[1]{\mathbb{#1}}

\newcommand{\wh}{\widehat}

\newcommand{\R}{\field{R}}

\newcommand{\var}{\textsc{var}}

\renewcommand{\tilde}{\widetilde}

\renewcommand{\Pr}{\mathbb{P}}
\newcommand{\hDelta}{\hat{\Delta}}

\newcommand{\ignore}[1]{}

\DeclareMathOperator*{\argmax}{argmax}
\DeclareMathOperator*{\argmin}{argmin}

\newtheorem{theorem}{Theorem}
\newtheorem{lemma}{Lemma}

\newcommand{\Prob}[1]{\mathbb{P}\left(#1\right)}
\newcommand{\set}[1]{\left\{ #1\right\}}

\title{On Learning to Rank Long Sequences\\ 
with Contextual Bandits}

\author{%
  Anirban~Santara \\
  Google~Research\\
  \texttt{santara@google.com} \\
   \And
   Claudio~Gentile\\
   Google~Research \\
   \texttt{cgentile@google.com} \\
   \And
   Gaurav~Aggarwal \\
   Google~Research \\
   \texttt{gagg@google.com} \\
   \AND
   Shuai~Li \\
   John~Hopcroft~Center \\
   Shanghai~Jiao~Tong~University \\
   \texttt{shuaili8@sjtu.edu.cn} \\

}

\begin{document}

\maketitle

\begin{abstract}
\vspace{-0.1in}
Motivated by problems of learning to rank long item sequences, we introduce a variant of the cascading bandit model that considers flexible length sequences with varying rewards and losses.  
We formulate two generative models for this problem within the generalized linear setting, and design and analyze upper confidence algorithms for it. Our analysis delivers tight regret bounds which, when specialized to vanilla cascading bandits, results in sharper guarantees than previously available in the literature. We evaluate our algorithms on a number of real-world datasets, and show significantly improved empirical performance as compared to known cascading bandit baselines.
\end{abstract}

\vspace{-0.1in}
\section{Introduction}
\vspace{-0.1in}
\looseness=-1
A well-known problem in content recommendation is the generation of {\em slates} of items whereby, given a set of available items and a limited number of available slots, the goal of the system is to come up with an ordered sequence of items to be arranged in the slots so as to best fulfil some goal, like improving the experience of the user at hand. Applications are ubiquitous, from web search to news recommendation, from computational advertising to web page content optimization. 
These are among the most prominent motivating applications behind the more abstract problem called {\em learning to rank}. 

The cascade model (e.g., \cite{chuklin2015click}) for learning-to-rank has emerged as a simple and effective way to model user behavior in a number of applications. In this model, the user scans the slate sequentially from top to bottom and clicks on the first item they find attractive, disregarding all subsequent items in the slate. The length of the slate may vary widely across applications, ranging from a few items in computational advertising to dozens in news recommendation to hundreds in web search. In these and many other dynamic domains, one has to deal with a near continuous stream of new items to be recommended, along with new users to be served. Out of the collected user feedback, and in the face of a constantly evolving content universe and set of targeted users, the learning system is expected to maintain over time a good mapping between user/item features and item rankings.

\looseness=-1
In order to encompass a variety of learning-to-rank applications for dynamic environments, we introduce a generalized version of the well-known cascading bandit model of \cite{kveton2015cascading}. Our model considers flexible sequence length with varying rewards and  losses. The problem is broadly described by position-dependent rewards $r_{j}$ and losses $\ell_j$. These parameters measure how well the ranking system is doing depending on the position $j$ of the first positive signal (e.g., the first click) as well as the potential loss associated with a sequence of $j$ negative signals. Since rewards are positive and losses are negative, and the two sequences are decreasing with $j$ (in particular, $\ell_j$ becomes more and more {\em negative} as $j$ increases), this model is intended to capture a natural trade-off in decision making. If we commit to a long sequence, we may increase our chance of success (positive reward), but also expose ourselves to the risk of a very negative loss if all signals on that sequence turn out to be negative.

This trade-off is typical in scenarios where each negative signal in the sequence is indeed a {\em cost} for the system. As a relevant example, suppose we want to deploy our ranking algorithm within a payment system (e.g., Stripe) where, at each round we process one transaction, and the goal is to find routes to fulfill the transaction. Here, each payment attempt comes with a cost for the system, the positive signal on a route corresponds to payment fulfillment through that route, while the negative signal corresponds to a payment failure. Every unsuccessful attempt reduces the net reward gathered by a subsequent success, and may translate into bigger losses if in the end the payment  is not fulfilled. This provides a classic use case of cascade models since we have to predict a ranked sequence of routes for the payment to be fulfilled with as few retries as possible. Note that the length of the ranked sequence can be large and flexible which further aligns this application to our setting. 

\looseness=-1
{\bf Our contribution.} In this paper, we describe two contextual upper confidence bandit algorithms for this problem, specifically focusing on the case of long ranked sequences. We analyze the two algorithms both theoretically and experimentally. Our theoretical analysis delivers tighter regret guarantees than previous investigations. In particular, we obtain a regret bound of the form $\sqrt{bT}$, where $T$ is the time horizon and $b$ is the length of the ranked sequences, as opposed to $b\sqrt{T}$ achieved by prior work in cascading bandits. We then validate our algorithms experimentally on well-known benchmark datasets, and show significantly improved performance as compared to the state-of-the-art algorithms.

{\bf Related work.}
The study of cascading bandit models for ranking problems has been initiated by \cite{kveton2015cascading}. The authors study the problem of learning to rank items on a fixed number of slots under the so-called cascade click model of user behavior. \cite{l+16,z+16,lz18} investigate large-scale variants where the reward of an item follows (generalized) linear structure. \cite{cheung2019thompson} gives an analysis for Thompson sampling. Cascading bandits have also been studied under more general click models, which can recover the standard cascade click model as well as other classical click models in the literature of online learning to rank (e.g., \cite{zoghi2017online,lattimore2018toprank,lls19}). \cite{li2019cascading} considers cascading bandits in non-stationary environments, and \cite{h+20} studies more comprehensive cascading models of user behavior that account for both position bias and diversity of recommendations. All these works consider the case of sequences with fixed length and, when specialized to the original cascading bandit model of \cite{kveton2015cascading} or generalized linear variants thereof, their analysis delivers a suboptimal dependence on the length of the sequence, which is a main theoretical concern in this paper. An in-context regret bound comparison to many of these works is carried out in Section \ref{s:independent}. Further related work is discussed in Appendix \ref{as:further}.

\vspace{-0.1in}
\section{Setting and Main Notation}
\vspace{-0.1in}
\looseness=-1
We formalize our problem of contextual bandits with long sequences as follows. Learning proceeds in a discrete sequence of {\em time steps} (or {\em rounds} or {\em trials}). 
At each time $t$, the learner processes a {\em transaction} having at its disposal a (finite) set of {\em actions} (or {\em items}) $A_t =\{x_{1,t}, x_{2,t}, \ldots, x_{k_t,t}\} \subseteq A = \{x \in \R^d\,:\, ||x||_2 \leq 1\}$, each action being described by a $d$-dimensional feature vector of (Euclidean) norm at most one.\footnote
{
This normalization is done for notational convenience only; any bounded action space would work here.
} 
Set $A_t$ is our {\em context} information at time $t$, while set $A$ is the universe of all possible actions. Collectively, $A_t$ may include information about the specific context in which learning is applied. In a payment scenario, this will typically include the transaction amount, the buyer and seller identities (or features), the credit card company identity (or features), etc. In a news recommendation problem this may include user features, news-of-the-day topic features, and so on.
Each action corresponds to an item available at time $t$. The learning problem is parameterized by a decreasing (or non-increasing) sequence of rewards $r_{1,t}, r_{2,t}, \ldots $
and a decreasing (or non-increasing) sequence of losses $\ell_{0,t}, \ell_{1,t}, \ell_{2,t}, \ldots $, where 
$$
1 \geq r_{1,t} \geq r_{2,t} \ge \ldots > 0~ \qquad and \qquad 0 > \ell_{0,t} \geq \ell_{1,t} \geq \ell_{2,t} \ge \ldots > -1~.
$$ 

\looseness=-1
The rewards are positive, while the losses are negative.
The dependence on $t$ of these quantities emphasizes the potential dependence of these values on the current context. For instance, in the payment scenario, $r_{i,t}$ is often proportional to the amount of the current transaction.
Moreover, to set the scale of these parameters, we shall assume throughout that $r_{i,t} \in [0,1]$ and $\ell_{i,t} \in [-1,0]$ for all $i$ and $t$.
Finally, each transaction may be accompanied by a {\em budget} value $b_t$ that bounds from above the number of allowed retries, as defined next.

In round $t$, the algorithm is compelled to play an ordered sequence of actions \ 
$
J_t = \langle x_{j_{1,t}}, x_{j_{2,t}}, \ldots x_{j_{s_t,t}}\rangle~,
$ 
where each component vector $x_{j_{i,t}}$ is taken from $A_t$. We call $J_t$ a {\em retry sequence} or simply a {\em sequence}\footnote
{
A sequence might have repeated actions, but for simplicity we assume here each component of $J_t$ is distinct.
}.
The set of all such sequences $J_t$ corresponds to the {\em action space} available to the learner at time $t$. Notice that the length $s_t$ of $J_t$ is part of the action selected by the learner (that is, the algorithm has to decide the length of the sequence as well). This length $s_t$ determines the number of retries on the transaction at time $t$. $J_t$ can also be empty; in such a case we have $s_t =0$ and write $J_t = \langle\rangle$. The budget constraint $b_t$ requires $s_t$ to satisfy $s_t \leq b_t$. In general, $b_t$ may depend on time, and there are practical scenarios where this is indeed advisable, e.g., a payment system where the number of attempts depends on the transaction amount.

\begin{figure}
    \centering
    \includegraphics[width=0.6\textwidth]{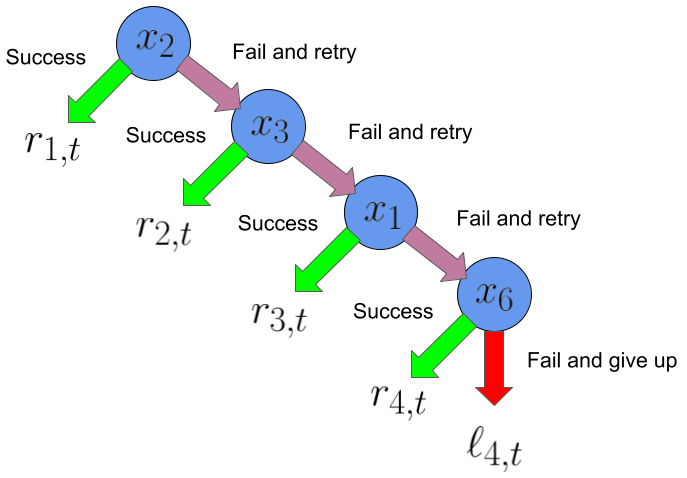}
    \caption{An illustration of the action space. Suppose at time $t$ the algorithm can select among items in $A_t = \{x_1, x_2, \ldots, x_{10}\}$, and that $b_t = 5$. In this example $J_t = \langle x_2,x_3,x_1,x_6\rangle$. If action $x_2$ succeeds, we gather $r_{1,t}$, while if $x_2$ fails, $x_3$, $x_1$, and $x_6$ are tried in turn. If all these actions fail, we incur the loss $\ell_{4,t}$. Also, notice that if $x_3$ was successful, we do not actually see whether $x_{1}$ and $x_{6}$ would have been successful or not.}
    \label{f:1}
\end{figure}

Sequence $J_t$ has associated rewards and losses as detailed next. Upon committing to $J_t$, if $J_t = \langle\rangle$ we simply suffer loss (or negative reward) $\ell_{0,t}$ and go to the next round. Otherwise, the first item $x_{j_{1,t}}$ is attempted. If $x_{j_{1,t}}$ is successful we gather reward $r_{1,t}$ and stop, going to the next round. If $x_{j_{1,t}}$ is unsuccessful, $x_{j_{2,t}}$ is attempted. If $x_{j_{2,t}}$ is successful we gather reward $r_{2,t}$ and again stop.  In this  way, finally, $x_{j_{s_t,t}}$ is attempted. If $x_{j_{s_t,t}}$ is successful we gather reward $r_{s_t,t}$ and stop. Otherwise, we ``give up" and incur loss $\ell_{s_t,t}$. A pictorial illustration is given in Figure \ref{f:1}.

\looseness=-1
The effort behind this parametrization for rewards and losses is to capture the tension between a potentially small reward of a successful late retry and a potentially small loss incurred by an early give up. On one hand, the earlier is the success in a sequence $J_t$ the higher the reward is likely to be. On the other, the later we give up (after many unsuccessful attempts) the higher is the loss we incur.

For simplicity, in this model rewards and losses incurred at time $t$ only depend on the position of the items in sequence $J_t$, rather than the actually played item in that position.
Also, upon processing the transaction at time $t$, the algorithm has to commit to the entire sequence $J_t$, that is, this sequence cannot be changed on the fly based on partial observations we are gathering on that sequence.\footnote
{
This is typically the case when the system is serving ranked content to (human) users.
} 
So, this is indeed a (parametric) cascading bandit model.

\looseness=-1
After playing $J_t$ at time $t$, the algorithm observes the reward associated with $J_t$, which is generated as follows. Let the {\em outcome} vector $Y_t$ be a Boolean vector $Y_t = (y_{1,t},\ldots, y_{|A_t|,t}) \in \{0,1\}^{|A_t|}$. Then we can define the reward $R_t(J_t,Y_t)$ of sequence $J_t$ at time $t$ (i.e., on the transaction occurring at time $t$) w.r.t. outcome $Y_t$ as follows (for ease of notation, we drop subscript $t$ and leave the dependence on $A_t$ implicit):
\begin{equation}\label{e:reward}
R(J,Y) = 
\begin{cases}
r_1 y_{j_{1}} 
+ \ldots +
r_{s} y_{j_{s}} \prod_{i=1}^{s-1}(1- y_{j_{i}}) + \ell_{s}  \prod_{i=1}^{s}(1- y_{j_{i}})
&{\mbox{if $J \neq \langle\rangle$}}\\
\ell_0 &{\mbox{otherwise}}~.
\end{cases}
\end{equation}
\looseness=-1
The above simply encodes the decision list exemplified by Figure \ref{f:1}, with the addition that if $J_t = \langle\rangle$ the algorithm decides to give up immediately, thereby incurring loss $\ell_{0,t}$, irrespective of the outcome vector $Y_t$.
As in standard cascading bandits, the algorithm does not observe the entire outcome vector $Y_t$, in fact, it specifically observes those components of $Y_t$ allowing to determine the actual value of reward $R_t(J_t,Y_t)$.

The outcome vector $Y_t$ is in turn generated according to the model described next.

\vspace{-0.1in}
\subsection{
Generative model}
\vspace{-0.1in}
%
Given the special form of the reward function, all we need to model are specific conditional probabilities. In order to properly define a generative model for $Y_t$, we start off by formally viewing $Y_t$ as a Boolean random vector $Y_t = (y_{1,t},\ldots, y_{|A_t|,t}) \in \{0,1\}^{|A_t|}$ with joint distribution $p_{Y_t}(A_t)$.
Notice that $Y_t$'s components need not be independent. The marginals and relevant conditional distributions of $p_{Y_t}(A_t)$ are defined as follows.
For simplicity, let $p(x_j)$ denote the (marginal) probability that item $x_j$ succeeds,

and 

\begin{equation}\label{e:condprob}
p(x_j\,|\,x_{i_1},\ldots,x_{i_k})
\end{equation}
be the probability that $x_j$ succeeds given that $x_{i_1}, \ldots, x_{i_k}$ have all failed. 

Once all conditional probabilities 
(\ref{e:condprob}) 
for all $x_j,x_{i_1},\ldots, x_{i_k}$ are available, we are automatically defining the generative process for the outcome $Y_t$ which is relevant to a sequence $J_t =  \langle x_{j_{1,t}},x_{j_{2,t}},\ldots,x_{j_{s_t,t}}\rangle$.

This is because, for the sake of computing $R_t(J_t,Y_t)$, the relevant events associated with $Y_t$ are those encoded by the strings 
\begin{equation}\label{e:strings}
\langle 1\rangle, \langle 0,1\rangle, \ldots, \langle \underbrace{0,\ldots,0}_{s_t-1\ zeroes},1\rangle, \langle \underbrace{0,\ldots,0}_{s_t\ zeroes}\rangle~,
\end{equation}
where the order of components within each string is determined by $J_t$, and,
\begin{align*}
\Prob{\langle \underbrace{0,\ldots,0}_{k\ zeroes},1\rangle} 
&= \prod_{i = 1}^{k-1} \Bigl(1-p(x_{j_{i,t}}\,|\,x_{j_{1,t}}, \ldots, x_{j_{i-1,t}})\Bigl)
\times\,p(x_{j_{k,t}}\,|\,x_{j_{1,t}}, \ldots, x_{j_{k-1,t}})~,\\[-2mm]
&\hspace{3.2in}{\mbox{for $k = 0,\ldots, s_t-1$,}}\\[-2mm]
\Prob{\langle \underbrace{0,\ldots,0}_{s_t\ zeroes}\rangle} 
&= \prod_{i = 1}^{k} \Bigl(1-p(x_{j_{i,t}}\,|\,x_{j_{1,t}}, \ldots, x_{j_{i-1,t}})\Bigl)~.
\end{align*}
We will soon give (\ref{e:condprob}) a parametric form. For the moment, observe that, based on the above generative model, we can define the expected reward $\E_{Y_t} [R_t(J_t,Y_t)]$ of $J_t$ on $A_t$ w.r.t. the random draw of $Y_t$. Specifically, if we take an expectation of (\ref{e:reward}) we obtain (we again drop subscript $t$ for readability):
\vspace{-0.05in}
\begin{align}\label{e:expregr}
\E_{Y} [R(J,Y)] 
= r_1\, p(x_{j_1}) 
+ \ldots
&\ + r_{s}\, p(x_{j_s}\,|\,x_{j_1},\ldots,x_{j_{s-1}}) \prod_{i=1}^{s-1}\Bigl(1- p(x_{j_i}\,|\,x_{j_1},\ldots,x_{j_{i-1}})\Bigl)\notag\\[-2mm] 
&\ + \ell_{s}\,  \prod_{i=1}^{s}\Bigl(1- p(x_{j_i}\,|\,x_{j_1},\ldots,x_{j_{i-1}})\Bigl)~,
\end{align}
and $\E_{Y} [R(J,Y)] = \ell_0$ if $J = \langle\rangle$.
The involved conditional probabilities (\ref{e:condprob}) are the only ones that matter in computing the expected reward $\E_{Y} [R(J,Y)]$. Moreover, the expected reward can be either positive or negative, due to the fact that the last term is negative. 

For a given pair $(A_t,b_t)$, a natural benchmark to compare to is the {\em Bayes optimal} sequence $J^*_t = \langle x_{j^*_{1,t}}, x_{j^*_{2,t}}, \ldots, x_{j^*_{s^*_t,t}} \rangle$, that is, the sequence $J_t$ that maximizes $\E_{Y_t}[R_t(J_t,Y_t)]$ over all possible sequences built on $A_t$, of length at most $b_t$. Recall that $J^*_t$ is computed by knowing beforehand all probabilities (\ref{e:condprob}) for all candidate sequences $J_t$. Consequently, we define the

time-$t$ (pseudo) {\em regret} of an algorithm that commits to $J_t$ on $A_t$ as
\(
\E_{Y_t}[R_t(J^*_t,Y_t)] - \E_{Y_t}[R_t(J_t,Y_t)]~,
\)
and its cumulative regret over $T$ rounds on the sequence of pairs $(A_1,b_1),(A_2,b_2) \ldots, (A_T,b_T)$ as
\[
\sum_{t=1}^T \E_{Y_t}[R_t(J^*_t,Y_t)] - \E_{Y_t}[R_t(J_t,Y_t)]~.
\]
\looseness=-1
Our goal is to make the above quantity as small as possible (with high probability).
Next, we formulate a parametric model for the  conditional probabilities (\ref{e:condprob}), and show: (i) how to compute $J^*_t$, and (ii) how to define the contextual bandit algorithms that determines $J_t$ so as to make the cumulative regret small.

\vspace{-0.1in}
\subsection{Parametric model} 
\label{ss:param}
\vspace{-0.1in}

Given our universe of actions $A=\{ x\in \R^d\,:\, ||x||_2 \leq 1\}$, we associate each item $x$ with a so-called coverage vector $c(x) = (c_{1}(x),\ldots, c_{d'}(x)) \in [0,1]^{d'}$, where $d'$ is the dimensionality of a latent space of {\em topics}.\footnote
{
Such coverage vectors can be obtained based on domain knowledge. E.g., they may be obtained as a latent probability distribution after training a Gaussian Mixture Model where the $d'$ Gaussian centroids represent the latent topics, and $c_{i}(x)$ is the probability that $x$ belongs to topic $i$ according to the mixture model. This is essentially what we do in our experiments in Section \ref{s:exp}.
}
The coverage $c_i(A')$ of a (finite) set $A' \subseteq A$ of items on topic $i$ is a monotone and sub-modular function on sets, e.g., $c_i(A') = 1 - \prod_{x \in A'} (1 - c_{i}(x))$, with $c_i(\emptyset) = 0$. Here we slightly abuse the notation and set $c_i(x) = c_i(\{x\})$. Following, e.g., \cite{yg11,h+20}, we then define the $d'$-dimensional vector $c'(x_j \mid x_{i_1}, \ldots, x_{i_{k}})$ of coverage differences, whose $i$-th component is
\[
c_i(\set{x_{i_1}, \ldots, x_{i_{k}}, x_j}) - c_i(\set{x_{i_1}, \ldots, x_{i_{k}}})\in [0,1]~.
\]
Since such vectors have only positive components, we shift them to their center so as both positive and negative components exist, and then divide by a constant that makes their norm at most $1$. For instance, we may set ${\bar c_i}(x_j \mid x_{i_1}, \ldots, x_{i_{k}}) = \frac{1}{\sqrt{d'}} \bigl(2c'_i(x_j \mid x_{i_1}, \ldots, x_{i_{k}}) - 1\bigl)$ to be the $i$-th component of the transformed vector ${\bar c}(\{x_{i_1},\ldots, x_{i_k}\})$ of coverage differences . 

Our parametric model is represented by a $d'$-dimensional vector $u \in \R^{d'}$ with the link function\footnote
{
As the reader can easily see, the content of this paper can be seamlessly extended to more general link functions (see, e.g., the treatment in \cite{DBLP:journals/corr/abs-1207-0166}) but, for simplicity of presentation, we restrict to the sigmoidal link.
} 
$\sigma\,:\,\R\rightarrow [0,1]$, $\sigma(z) = \frac{\exp{(z)}}{1+\exp{(z)}}$. Specifically we set the conditional probability as
\begin{equation}\label{e:dependent_model}
    p(x_j \mid x_{i_1}, \ldots, x_{i_k}) = \sigma({\bar c}(x_j \mid x_{i_1}, \ldots, x_{i_k})^\top u)~.
\end{equation}
Hence the marginal probabilities $p(x)$ and conditional probabilities $p(x_j \mid x_{i_1}, \ldots, x_{i_k})$ are encoded as generalized linear functions with unknown parameter vector $u$.
The idea behind this model is that if the additional topic-wise diversity brought up by $x_j$ as compared to the already selected $x_{i_1}, \ldots, x_{i_k}$ is relevant w.r.t. the weight vector $u$, then the probability that $x_j$ is successful given that $x_{i_1}, \ldots, x_{i_k}$ has failed should be large. The opposite happens if the additional diversity contributed by $x_j$ is indifferent w.r.t. $u$.

We now separate two cases: (i) the independent outcome case, where only marginal probabilities $p(x)$ are needed, and (ii) the more general dependent outcome case, where also the conditional probabilities $p(x_j\,|\,x_{i_1},\ldots,x_{i_k})$ have to be considered. As we will see in the sequel, (ii) reduces to (i), up to the computation of $J^*_t$. For the independent case we can simply set ${\bar c}(x\,|\, x_1,\ldots, x_k) = x$, for all $x, x_1,\ldots, x_k$,
and $d'=d$ to save notations, which makes $p(x) = \sigma(u^\top x)$.

\vspace{-0.1in}
\section{Independent Outcomes}\label{s:independent}
\vspace{-0.1in}
This is the simplest possible setting where the Boolean vector $Y_t$ has independent components. In this case, in (\ref{e:condprob}) we have $p(x_j\,|\,x_{i_1},\ldots,x_{i_k}) = p(x_j)$ for all $x_{i_1}, x_{i_2}, \ldots, x_{i_k}$, and $x_j$. Hence there is no reason to model conditional probabilities, and we restrict to modeling $p(x) = \sigma(u^\top x)$. Moreover, in this case, Bayes is formulated only by means of marginal probabilities $p(x_i)$, and reduce to sorting items in $A_t$ in decreasing order of $p(x_j)$ and stopping when a suitable condition is met. We now claim that, in this specific case, the Bayes optimal sequence $J^*$ can be computed fairly easily.
Due to space limitations, all proofs are postponed to the appendix.

\begin{lemma}\label{l:bayesindep}
Let $p_Y(A) = \prod_{j=1}^{|A|} p(x_j)$, and $b$ be the budget length. Then $J^*$ can be computed as follows. Set
\(
s^* = \arg\max_{s = 0, 1,\ldots, b} E_{Y} [R(J^*_s,Y)]~,
\)
where $J^*_s = \langle x_{j^*_{1}},x_{j^*_{2}},\ldots,x_{j^*_{s}}\rangle$, $x_{j^*_{1}},x_{j^*_{2}},\ldots,x_{j^*_{s}}$ the items associated with the $s$ largest marginal probabilities $p(x_j)$, $x_j\in A$, sorted in non-increasing order. Then $J^* = J^*_{s^*}$,
with $J^* = \langle\rangle$ if $s^*=0$.
\end{lemma}

\setlength{\textfloatsep}{0.2cm}
\setlength{\floatsep}{0.2cm}
\begin{algorithm*}[t!]
{\bf Input:} Confidence level $\delta \in [0,1]$,  width parameter $D > 0$, maximal budget parameter $b>0$\\
{\bf Init:} $M_{0} = bI \in \R^{d\times d}$,\ $w_{1} = 0\in \R^{d}$, $c_1 = 1$ \\
{\bf For $t=1, 2,\ldots, T$}
\begin{enumerate}
\item Get:
\vspace{-0.05in}
\begin{itemize}
    \item Set of actions $A_t = \{x_{1,t},\ldots, x_{|A_t|,t}\} \subseteq \{  x \in \R^d\,:\,||x|| \leq 1\}$~,
    \vspace{-0.05in}
    \item budget $b_t \leq b$~;
\end{itemize}
\vspace{-0.05in}
\item For $x_j \in A_t$, set $\hDelta_{j,t} = x_j^\top w_{c_t}$~;
\vspace{-0.05in}
\item Compute $J_t$~:
\vspace{-0.05in}
\begin{itemize}
    \item Let ${\wh J_{t,s}} = \langle x_{\wh j_{t,1}}, \ldots, x_{\wh j_{t,s}} \rangle $ be made of the $s$ largest items in $A_t$ in non-increasing order of $\hp_{j,t}$, where\,:\ \ \
\begin{itemize}
\vspace{-0.05in}
\item
\(
\hp_{j,t} = \sigma(\hDelta_{j,t}+\epsilon_{j,t}),
\)
\vspace{-0.05in}
\item \(
\epsilon^2_{j,t} 
= 
(x_j^\top M^{-1}_{c_t-1} x_j)\,\,\alpha(b,d,T,\delta)~,
\)
~~with 
\vspace{-0.05in}
\begin{align*}
\alpha(b,d,T,\delta) = &2bD^2 
+ \left(\frac{c_{\sigma}}{c_{\sigma'}}\right)^2 d\log \left(1+ \frac{2}{b}\Bigl(T\,\frac{c_{\sigma}}{1-c_{\sigma}} + 4\log\frac{4(T+1)}{\delta}\Bigl) \right)\\
& + 2\left(12\left(\frac{c_{\sigma}}{c_{\sigma'}}\right)^2 + \frac{36(1+D)}{c_{\sigma'}} \right) \log \frac{2b(T+4)}{\delta}
+  20D^2\log\frac{2bd(T+1)}{\delta}
\end{align*}
\end{itemize}
\vspace{-0.05in}
\item Set ${\wh s_t} = \arg\max_{s=0,1,\ldots,b_t} {\wh \E_{Y_t}}[R({\wh J_{t,s}},Y_t)] $~,\\
with
\vspace{-0.05in}
\[
{\wh \E_{Y_t}}[R({\wh J_{t,s}},Y_t)] = 
\begin{cases}
E\left(\hDelta_{{\wh j_{t,1}},t}+\epsilon_{{\wh j_{t,1}},t},\ldots,\hDelta_{{\wh j_{t,s}},t}+\epsilon_{{\wh j_{t,s}},t}\right) &{\mbox{if $s \geq 1$}}\\
\ell_{0,t} &{\mbox{otherwise}}~,
\end{cases}
\]
where function $E(\cdot,\ldots,\cdot)$ is as in (\ref{e:ehat}) in Lemma \ref{l:monotonicity} (see Appendix \ref{as:analysis}) with $p(\cdot)$ therein replaced by $\sigma(\cdot)$~;

\item Finally, $J_t = {\wh J_{t,{\wh s_t}}}$~;
\end{itemize}
\vspace{-0.05in}
\item Observe feedback $Y_t \downarrow J_t = 
\begin{cases}
\langle y_{t,{\wh j_{t,1}}}, y_{t,{\wh j_{t,2}}}, \ldots, y_{t,{\wh j_{t,{\wh s'_t}}}}\rangle = \langle 0,\ldots, 0, 1\rangle, &{\mbox{for some ${\wh s'_t} \leq {\wh s_t}$\qquad or}}\\
\langle y_{t,{\wh j_{t,1}}}, y_{t,{\wh j_{t,2}}}, \ldots, y_{t,{\wh j_{t,{\wh s_t}}}}\rangle = \langle 0,\ldots, 0, 0\rangle
\end{cases}
$
\item {\bf For} $j = 1,\ldots, {\wh s_t}$ (in the order of occurrence of items in $J_t$)
update:
\[
M_{c_t+j-1} = M_{c_t+j-2} + |s_{j,t}| x_j x_j^\top,\qquad 
w_{c_t+j} = w'_{c_t+j-1} + \frac{1}{c_{\sigma'}} M^{-1}_{c_t+j-1}\nabla_{j,t}~,
\]
where
\vspace{-0.12in}
\[
s_{j,t} =
\begin{cases}
1 & {\mbox{If $y_{t,j}$ is observed and $y_{t,j}= 1$}}\\
-1 & {\mbox{If $y_{t,j}$ is observed and $y_{t,j}= 0$}}\\
0 & {\mbox{If $y_{t,j}$ is not observed}}~,
\end{cases}
\]
and
\(
\nabla_{j,t} = \sigma(-s_{j,t}\,\hDelta'_{j,t})\,s_{j,t}\,x_j~,
\)
where $\hDelta'_{j,t} = x_j^\top w'_{c_t+j-1}$\\ 
with
$$
w'_{c_t+j-1} = \arg\min_{w\,:\,-D \leq w^\top x_j \leq D} d_{c_t+j-2}(w,w_{c_t+j-1})~;
$$
\vspace{-0.1in}
\item $c_{t+1} \leftarrow c_t + {\wh s_t}$~.
\end{enumerate}

\caption{\label{f:2}The contextual bandit algorithm in the independent case. Here the link function $\sigma(\cdot)$ is $\sigma(x) = \frac{\exp(x)}{1+\exp(x)}$.}

\end{algorithm*}

The bandit algorithm corresponding to (or mimicking) the Bayes computation in Lemma \ref{l:bayesindep} is described in Algorithm \ref{f:2}. In this pseudo-code and elsewhere, we use the notation $Y_t \downarrow J_t$, henceforth called outcome {\em projected} onto the retry sequence, to denote the binary string of the form (\ref{e:strings}) which encodes the components of outcome vector $Y_t$ that are revealed by playing sequence $J_t$. 
Recall Figure \ref{f:1} for an example: If $Y_t = (0,0,1,1,0,1,1,0,0,1)$ and $J_t = \langle x_1,x_2,x_7,x_{10}\rangle$ we have $Y_t \downarrow J_t = \langle 0,0,1\rangle$, that is, playing $J_t$ when the outcome is $Y_t$ reveals the components of $Y_t$ in the order determined by $J_t$ up to the first 1 in $Y_t$. In this example, we observe the 1st, the 2nd, and the 7th component of $Y_t$. Notice, in particular, that we do not observe $Y_t$'s 10th component. 

\looseness=-1
Algorithm \ref{f:2} replaces the true marginal probabilities $p(x_j) = \sigma(u^\top x_j)$ with upper confidence estimations $\hp_{j,t} = \sigma(\hDelta_{j,t}+\epsilon_{j,t})$, and then mimics the Bayes optimal computation to determine $J_t$. The update rule is a second-order descent method on an appropriate loss function (logistic, in this case) associated with the link function $\sigma$. Notice that the items $x_j$ which do not occur in $Y_t \downarrow J_t$ have $s_{j,t} = 0$, hence they do not contribute to the update of $M_{t}$ or $w_{t}$. Yet, it is important to emphasize that $s_{i,t}$ can be zero (that is, the corresponding component $y_{t,i}$ is not observed) also due to the fact that an earlier item than $x_i$ in $J_t$ has been successful. The update of vector $w_{c_t+j-1} \rightarrow w_{c_t+j}$ is done by first projecting $w_{c_t+j-1}$ onto the set $\{w\in \R^d\,:\, |w^\top x_j| \leq D\}$ to obtain $w'_{c_t+j-1}$, and then computing a standard Newton step. The projection can be efficiently calculated in closed form (see Appendix \ref{as:analysis}).

\looseness=-1
A convenient way of viewing the way the algorithm works is as follows.
The time horizon is split into rounds $t = 1, 2, \ldots, T$, each round containing multiple update steps. At the beginning of round $t$, the algorithm commits to a sequence $J_t$ of length ${\wh s_t}$ using the weight vector $w'_{c_t}$ available at {\em the beginning} of that round. Then feedback sequence $Y_t \downarrow J_t$ of length ${\wh s'_t} \leq {\wh s_t}$ is observed and a sequence $j = 1, \ldots, {\wh s'_t}$ of updates are executed within round $t$. The remaining ${\wh s_t} - {\wh s'_t}$ are those corresponding to $s_{j,t} = 0$.
Notice that, unlike the cascading contextual bandit algorithms available in the literature (e.g., \cite{z+16,l+16,lz18,llz18,li19,lls19,h+20}), our Algorithm \ref{f:2} clearly tells apart through the update rule the actions in the sequence $J_t$ that have been observed to be failures ($s_{j,t} = -1$) and those that have not been observed at all ($s_{j,t} = 0$). As shown in the appendix, this richer update rule help us prove a sharper regret guarantee than those available in the literature. The next is the main result of this section.
\begin{theorem}\label{t:independent}
Assume there exists $D >0$ such that $u^\top x_j \in [-D,D]$ for all $x_j \in A$.\footnote
{
Notice that since we have assumed $||x_j||_2 \leq 1$ for all vectors $x_j$, we also have $||u||_2 \leq D$.
}
Let $c_{\sigma}$ and $c_{\sigma'}$ be two positive constants such that, for all $\Delta \in [-D,D]$ we have $0 < 1-c_{\sigma} \leq \sigma(\Delta) \leq c_{\sigma} < 1$ and $\sigma'(\Delta) \geq c_{\sigma'}$.
Then with probability at least $1-\delta$, with $\delta < 1/e$, the cumulative regret of Algorithm \ref{f:2} run with a link function $\sigma\,:\,\R \rightarrow [0,1]$ such that $\sigma'(\Delta) \leq z$ for all $\Delta \in \R$ satisfies
\begin{align*}
\sum_{t=1}^T \E_{Y_t} [R(J^*_t,Y_t)] - \E_{Y_t} [R(J_t,Y_t)]\leq
4z\sqrt{\frac{e\,c_{\sigma}}{1-c_{\sigma}}T \alpha(b,d,T,\delta) }\,d\log(1+T)~,
\end{align*}
where $\alpha(b,d,T,\delta)$ is the log factor
\begin{align*}
O\Biggl[bD^2 + \left(\frac{c_{\sigma}}{c_{\sigma'}}\right)^2\hspace{-0.05in}d\log \left(1+ \frac{1}{b}\Bigl(\frac{T\,c_{\sigma}}{1-c_{\sigma}} + \log\frac{T}{\delta}\Bigl) \right)
+ \left(\left(\frac{c_{\sigma}}{c_{\sigma'}}\right)^2 + \frac{1+D}{c_{\sigma'}} \right) \log \frac{bT}{\delta}
 + D^2\log\frac{bdT}{\delta}\Biggl]~,
\end{align*}
the big-oh hiding additive and multiplicative constants independent of $T$, $d$, $b$, $D$, $\delta$, $c_\sigma$, and $c_{\sigma'}$.
\end{theorem}
\begin{remark}
Here and throughout, since $\sigma(x) =\frac{\exp{(x)}}{1+\exp{(x)}}$, we have $c_{\sigma} = \frac{e^D}{1+e^D}$ (so that $\frac{c_{\sigma}}{1-c_{\sigma}} = e^D$), $c_{\sigma'} = e^{-D}/(1+e^{-D})^2 \geq e^{-D}/4$, and $z = 1$. The dependence on $e^D$ is common to all logistic bandit bounds,\footnote
{
This actually applies only to the so-called frequentist regret bounds, which are the ones considered here. Switching to a Bayesian regret guarantee allows one to give bounds which, under some conditions, are independent of $D$ -- see~\cite{dmv19}. Staying within the realm of frequentist guarantees, it might be possible to improve Theorem \ref{t:independent} by following the more refined self-concordant analysis contained in \cite{DBLP:journals/corr/abs-2002-07530}. This analysis allows one to move the multiplicative dependence on $e^D$ from $\sqrt{T}$ to a lower order term in $T$.
}
and is due to the nonlinear shape of $\sigma(\cdot)$ (see, e.g., \cite{f+10,DBLP:journals/corr/abs-1207-0166,zhang2016online,li2017provably,DBLP:journals/corr/abs-2002-07530}, where it takes the form of  an upper bound on $1/\sigma'(\cdot)$). Also notice that $D$ is meant to be a constant here. 
As for the dependence on the sequence length $b$, our bound has the form ${\tilde O}(\sqrt{bT})$. Yet, we would like to emphasize that if we are willing to pay an extra {\em additive} term of the form $e^b$ in the regret guarantee, there is a simple way to obtain a bound of the form ${\tilde O}(e^b + \sqrt{T \log b})$ through a more careful tuning of $b$ in Algorithm \ref{f:2}. Specifically, following \cite{li19}, we can set
\vspace{-0.05in}
\begin{align*}
b = \argmin_{b \geq \max_t b_t} \Biggl(bD^2 + \left(\frac{c_{\sigma}}{c_{\sigma'}}\right)^2 d\log \left(1+ \frac{1}{b}\Bigl(T\,e^D + \log\frac{T}{\delta}\Bigl) \right)\Biggl)
\end{align*}
to achieve the claimed guarantee.
\end{remark}
\vspace{-0.07in}
\looseness=-1
\noindent{\bf Regret bound comparison.}
Many papers have tackled the problem of cascading bandits with contextual information, some of them adopting a linear model assumption~(e.g., \cite{z+16,l+16,lls19,h+20}), others a generalized linear model assumption~(e.g., \cite{lz18,llz18,li19}). Most of these papers have been chiefly motivated by learning-to-rank tasks applied to recommendation problems. 
Our usage of cascading bandits may be motivated by widely different application domains, where the sequence $J_t$ can potentially be far longer than the ranked list of items typically served to the user of an online content provider. So, we are interested in both the dependence on the time horizon $T$ and the maximal length $b$.
Our bound of the form $\sqrt{bT}$ improves on past results in  contextual cascading bandits, where the dependence on $b$ is either of the form $b\sqrt{T}$ (\cite{z+16,lls19,h+20}) or of the form $b\sqrt{bT}$ (\cite{llz18}) or of the form $e^b+\sqrt{bT}$ (\cite{lz18,li19}) or even of the form $\frac{1}{p^*}\sqrt{bT}$ (\cite{l+16}), where $p^*$ is the smallest probability of any sequence of length $b$, which can easily be exponentially small in $b$, even in the case of independent outcomes considered here.

\vspace{-0.14in}
\section{Dependent Outcomes}\label{ss:dependent}
\vspace{-0.1in}
Starting from the parametric model of Section \ref{ss:param}, we can write the conditional probabilities as
\begin{eqnarray*}
p(x_{j_{k+1}}\,|\, x_{j_1},\ldots, x_{j_k})
&=& \sigma(\Delta_{j_1,\ldots, j_k, j_{k+1}})~,
\end{eqnarray*}
where
\(
\Delta_{j_1} = c(x_{j_1})^\top u
\)\ 
and
\(
\Delta_{j_1,\ldots, j_k, j_{k+1}} = c(x_{j_{k+1}} \mid x_{j_1}, \ldots, x_{j_k})^\top u~,
\) 
for all $k \geq 1$.
With this notation, the expected regret (\ref{e:expregr}) can be written as
\begin{equation}\label{e:expregrdepdendent}
\E_{Y}[R(J,Y)]=
\begin{cases}
E(\Delta_{j_1}, \ldots, \Delta_{j_1, \ldots, j_s}) &{\mbox{if $J \neq \langle\rangle$ }}\\
\ell_0 &{\mbox{otherwise}}~,
\end{cases}
\end{equation}
where $E(\cdot,\ldots, \cdot)$ is defined in (\ref{e:ehat}) (see Appendix \ref{as:analysis}) with $p(\cdot)$ therein replaced by $\sigma(\cdot)$.

\looseness=-1
The algorithm operating with the above generative model is an adaptation of the one we presented for the independent case. The main difference here is that we use conditional probabilities computed from  coverage difference vectors. Notice that calculating $J^*$ may be computationally intractable. Yet, having at our disposal an oracle that maximizes (\ref{e:expregrdepdendent}) over $J$, we could clearly carry out a formal regret analysis similar to the one in Theorem \ref{t:independent}.
As in \cite{h+20}, we resort to a greedy algorithm to reduce the computational complexity. 
Specifically, we give an order over all candidate items based on their coverage difference vectors $c(\cdot \mid x_{\wh j_{t,1}}, \ldots, x_{\wh j_{t,k-1}})$ w.r.t. the already listed items. Then the empirical mean and upper confidence levels are computed based on these difference vectors, while the length of the sequence is chosen based on a search over all possible length values with the computed upper confidence levels.

Below we describe a simple greedy algorithm (henceforth called {\sc Greedy}) operating on true probabilities $p(x_{j_{k+1}}\,|\, x_{j_1},\ldots, x_{j_k})$, and give the pseudocode of its bandit counterpart in Appendix \ref{as:dependent}. This bandit {\sc Greedy} will be tested in our experimental comparison in Section \ref{s:exp}.

For convenience, we drop subscript $t$.
On the set of available actions $A$, the algorithm builds sequence $J_s = \langle x_{j_1}, x_{j_2}, \ldots, x_{j_{s}}\rangle$
of length $s \leq b_t$
as follows. 
For $k = 1,\ldots, s$, append
\begin{equation}\label{eq:greedy}
x_{j_k} = \arg\max_{x \in A\setminus \{x_{j_1}, \ldots, x_{j_{k-1}}\}} p(x\,|\,x_{j_1}, \ldots, x_{j_{k-1}})~,
\end{equation}
to $\langle x_{j_1}, x_{j_2}, \ldots, x_{j_{k-1}}\rangle$. 

As for the analysis, let $J_t^\ast$, the Bayes optimal sequence at time $t$, have length $s_t^\ast$. Then it can be proven that the greedy algorithm gives an approximation ratio $0 < \gamma(s_t^\ast) < 1$, with some mild assumptions on rewards $r_i$ and losses $\ell_i$. Such a ratio is unavoidable since the optimal offline solution (when the true probabilities are known) is computationally intractable. Like previous work on combinatorial multi-armed bandits with an approximation oracle (e.g., \cite{h+20}), we also consider the {\em scaled} cumulative regret, where one-time regret is defined as
\begin{align}\label{e:onetimescaled}
\E_{Y} [\gamma(s_t^\ast) R(J_t^*,Y)] - \E_{Y} [R(J_t, Y)] \,.
\end{align}
Then by a result similar to Lemma \ref{l:onetimeregret} for the independent case, we can derive a regret bound of the form $\sqrt{\alpha(b,d',T,\delta)T}\,d'\log T$.
The detailed derivation of $\gamma(s_t^\ast)$ and the proof of the key lemmas are given in Appendix \ref{as:dependent}.

\vspace{-0.1in}
\section{Experiments}\label{s:exp}
\vspace{-0.1in}
\looseness=-1
In order to demonstrate the efficacy of the proposed algorithms, we present our experimental results on ranking tasks defined on the Million Songs ~\cite{Bertin-Mahieux2011}, Yelp~\cite{yelp}, MovieLens-25M~\cite{harper2015movielens}, and MNIST~\cite{mnist} datasets. We compare our algorithms to exploration-exploitation baselines in the cascading bandits literature,\footnote
{
We believe this set of baselines are collectively a good pool of representatives of the relevant literature. Notice, in particular, that we do not compare to traditional learning to rank methods that do not rely on exploration/exploitation, since the partial information structure of our problem would make this comparison somewhat questionable.
}
specifically to the CascadeUCB1 algorithm of~\cite{z+16} (called ``C-UCB1'' later on), the GL-CDCM algorithm of \cite{liu2018contextual} which relies on a generalized linear model with the original Maximum Likelihood Estimator (MLE) as in \cite{f+10}, an $\epsilon$-greedy version of our Algorithm \ref{f:2} (called ``Eps'' later on), and a purely random policy 
(called ``Rand'' later on),

\looseness=-1
{\bf Datasets and preprocessing. }
We describe the pre-processing steps used for the MovieLens-25M dataset. 
The Million Songs and Yelp datasets have been treated using similar steps. 
MovieLens-25M contains ratings of $59,047$ movies by $162,541$ users, and is popularly studied in the recommendation system literature. We sample $10,000$ movies at random and calculate the singular value decomposition (SVD) of the corresponding $162,541 \times 10,000$ ratings matrix into $10$ principal components. The projection matrices from the SVD are used to compute embeddings of dimension $d=10$ for the remaining $49,047$ movies for training the bandit algorithms. The embeddings are normalized to unit $L_2$-norm and the dataset is shuffled randomly. In every round of bandit learning, the algorithm is presented with a non-overlapping chunk of movies as arms ($A_t$). The chunk size is $100$ (except for the last one, which is of size 47).
The rate of success of an arm is decided by the mean rating received by the corresponding movie in the dataset. This mean rating is normalized by first re-centering through its median value in the dataset, and then converting to a probability by passing through a sigmoidal function.
As mentioned in Section \ref{ss:param}, for the dependent algorithm the 49,047 SVD-projected $d$-dimensional vectors have been used to compute coverage vectors through a Gaussian Mixture Model (GMM) with $d'$ centroids.
As for MNIST, this is a multi-class classification dataset. We designed $10$ ranking tasks out of it, one for each of the $10$ classes in the dataset. Each task has one class as the ``pivot-class''. The algorithm must rank a collection of samples to have an item of the pivot-class (if present) as high up in the list as possible. Further details on pre-processing can be found in the appendix. 

{\bf Scenarios.} 
\looseness=-1
We study two reward/loss scenarios. The first one, which we call \emph{``Vanilla''}, is designed to reproduce the standard scenario studied in the traditional cascading bandit literature: $r_{j,t} = 1$, for all $t$ and $j = 1, 2, \dots, b_t$, and $\ell_{j,t} = 0$, for all $t$ and $j = 0, 1, \dots, b_t$. The second scenario, called \emph{``Exponential''} is comprised of exponentially decaying rewards and losses, and is designed to incentivize early success:

            \(
                r_{j,t} = \frac{1}{2^{j-1}}, \mbox{ for all $t$ and } j = 1, 2, \dots, b_t, \mbox{ and } 
                \ell_{j,t} = \frac{4}{5} \times \frac{1}{2^{j}} - 1,  \mbox{ for all $t$ and } j = 0, 1, \dots, b_t.
            \)

Notice that in the exponential scenario $r_{1,t}=1$ and
$\ell_{0,t} = -0.2$. The exponential scenario captures the true essence of the proposed models since it remains sensitive to early success even for larger budgets.

\looseness=-1
{\bf Tuning of Hyperparameters.}
We run a fine grid-search over the space of hyperparameters of each algorithm and only report the results corresponding to the combination of hyperparameters that obtains the largest final cumulative reward. We search the value of learning rate\footnote
{
A learning rate is introduced in the Newton step $w_{c_t+j} = w'_{c_t+j-1} + \frac{1}{c_{\sigma'}} M^{-1}_{c_t+j-1}\nabla_{j,t}$ of our algorithms so as to replace the ``theoretically-motivated" (and overly conservative) factor $\frac{1}{c_{\sigma'}}$.
} 
in the range $1.0-100.0$, UCB exploration parameter $\alpha = \alpha(b,d,T,\delta)$ or $\alpha = \alpha(b,d',T,\delta)$ on a logarithmic scale between $10^{-9}-10.0$, $\epsilon$ in $\epsilon$-greedy in the range $0.01-0.5$, L$2$ regularization weight $\lambda$ in our implementation of the GL-CDCM baseline \cite{liu2018contextual} on a logarithmic scale between $10^{-7}-10^{3}$ and the number $d'$ of latent components for the proposed dependent algorithm between $3$ and $30$.

\begin{figure}[t!]
    \centering
    \includegraphics[width=0.255\textwidth]{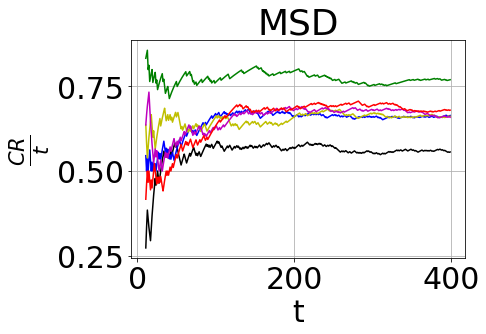} \hspace{-0.10in}
    \includegraphics[width=0.255\textwidth]{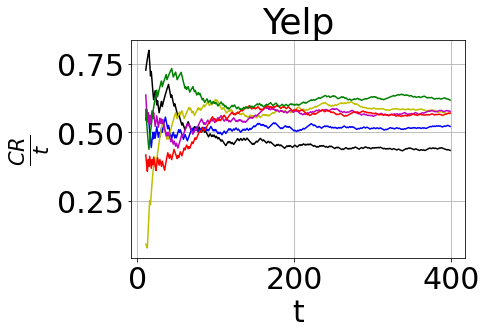} \hspace{-0.10in}
    \includegraphics[width=0.24\textwidth]{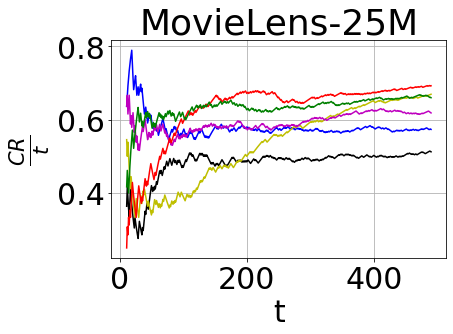} \hspace{-0.10in}
    \includegraphics[width=0.24\textwidth]{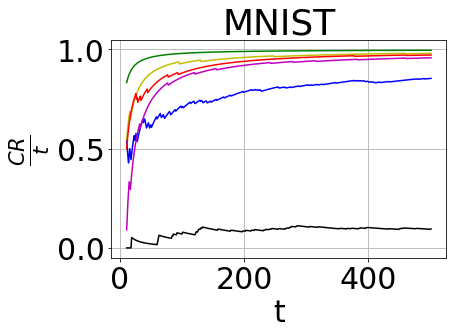}\\
    ~~\includegraphics[width=0.255\textwidth]{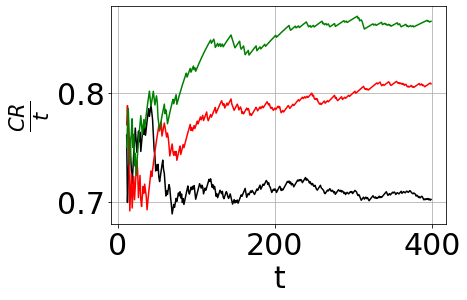} \hspace{-0.11in}
    \includegraphics[width=0.255\textwidth]{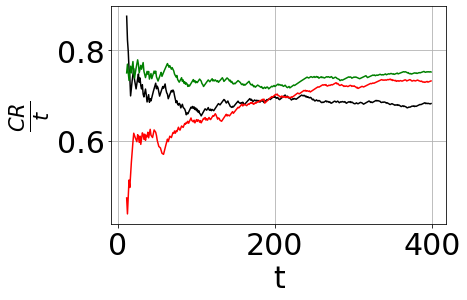} \hspace{-0.11in}
    \includegraphics[width=0.24\textwidth]{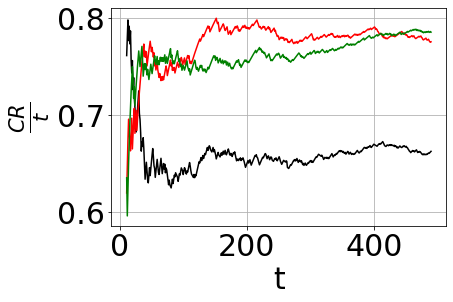} \hspace{-0.11in}
    \includegraphics[width=0.24\textwidth]{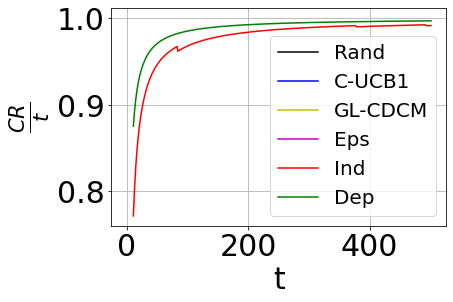}
    \vspace{-0.15in}
    \caption{Average cumulative reward $CR(t)/t$ as a function of $t = 1,\ldots,T$ for the various algorithms on Million Songs Dataset (MSD), Yelp,  MovieLens-25M, and MNIST with pivot-class $0$, respectively.
    Vanilla scenarios are on the top row, exponential scenarios on the bottom row. $b_t = 1$ for vanilla and $b_t=10$ for exponential. In the bottom right plot, Rand is not included for better visibility. The proposed Dependent (``Dep'') algorithm performs best across the datasets, with an exception of MovieLens-25M, where the proposed Independent (``Ind'') algorithm performs slightly better.
   }
    \label{fig:movielens-cr/time}
\end{figure}

\looseness=-1
{\bf Results.}
Figure \ref{fig:movielens-cr/time}
contains an experimental comparison among all algorithms.
We evaluate the algorithms in terms of their time-averaged Cumulative Reward ($CR$) obtained over all rounds of training by computing, for each algorithm, the fraction of reward/loss units accumulated per time step, up to time $t$, for $t = 1,\ldots, T$. 
If a given dataset has $T$ chunks then each algorithm is trained for exactly $T$ rounds.
Figure \ref{fig:movielens-cr/time} shows the variation of $\frac{CR(t)}{t}$ over rounds of training for two of the scenarios that incentivize early success.
In the exponential scenario, we restrict to comparing Rand, Ind, and Dep, since the other baselines are not designed to cope with it.
Notice that, since the vanilla scenario does not distinguish between early and late successes in the sequence, for larger values of $b_t$ the performances of all algorithms become indistinguishable from one another.
For lower values of $b_t$ in the vanilla scenario and all values of $b_t$ in the exponential scenario, achieving higher $CR$ is synonymous of early success, and we observe that the proposed dependent algorithm (``Dep'') outperforms the other algorithms in these scenarios, an exception being MovieLens-25M, where the proposed Independent (``Ind'') algorithm performs slightly better. GL-CDCM turns out to be a strong competitor, often at par with Ind, though it should be emphasized that the MLE estimation in GL-CDCM makes its running time far higher than that of Ind and Dep.
Further experimental results are provided in the appendix, where similar trends as those reported here can be observed.

\vspace{-0.1in}
\section{Conclusions}
\vspace{-0.1in}
We have introduced a cascading bandit model with flexible sequences and varying rewards and losses. The model is specifically focused on learning-to-rank applications, like web search or payment systems, where the item sequence can be significantly long. We have analyzed two algorithms with improved regret guarantees, and have empirically demonstrated their competitiveness against standard baselines on a number of well-known real-world benchmark datasets.

\bibliographystyle{plainnat}
\bibliography{biblio}



\newpage

\onecolumn

\appendix
\section{Appendix}\label{as:analysis}

The following lemma is of preliminary importance. It delivers a monotonicity property showing that the upper confidence scheme adopted in Algorithm \ref{f:2} below is properly defined, but it also serves in the proof of  subsequent lemmas.
%
\begin{lemma}\label{l:monotonicity}
For constants 
$r_{1} \geq r_{2} \ldots \geq r_s > 0$, $\ell_{s} < 0$, and a differentiable function $p:\, \R \rightarrow [0,1]$ which is monotonically increasing, the function $E\,:\, \R^s \rightarrow \R$ defined as
\begin{align}
E&(\Delta_1,\Delta_2,\ldots, \Delta_s)\notag\\ 
&= r_1\, p(\Delta_1) + r_2\, p(\Delta_2) (1- p_(\Delta_1)) +
\ldots + r_{s}\, p(\Delta_s) \prod_{i=1}^{s-1}(1- p(\Delta_i)) + \ell_{s}\,  \prod_{i=1}^{s}(1- p(\Delta_i))\label{e:ehat}
\end{align}
enjoys the following properties:
\begin{enumerate}
\item $E$ is non-decreasing in each individual variable $\Delta_i$. 
\item If, in addition, $r_i \in [0,1]$, for $i = 1,\ldots, s$, $\ell_s \in [-1,0]$, and $\frac{dp(\Delta)}{d\Delta} \leq z$ for all $\Delta \in \R$, then
\(
\frac{\partial E(\Delta_1,\ldots, \Delta_s)}{\partial \Delta_i} \leq z(r_i-\ell_s) \leq 2z~
\)
holds for all $\Delta_1,\ldots, \Delta_s \in \R$, and $i$.
\item Under the same assumption as in item 2 above, 
\[
\frac{\partial E(\Delta_{1},\ldots, \Delta_s)}{\partial \Delta_k} \leq 2z\,\prod_{j=1}^{k-1}(1-p(\Delta_j))~.
\]
\end{enumerate}
\end{lemma}
\begin{proof}
Define, for $k = 1,\ldots, s$,
\begin{align*}
E_k &= E_k(\Delta_k,\Delta_{k+1},\ldots, \Delta_s)\\ 
&= r_k\, p(\Delta_k) + r_{k+1}\, p(\Delta_{k+1}) (1- p(\Delta_k)) + \ldots  + r_{s}\, p(\Delta_s) \prod_{i=k}^{s-1}(1- p(\Delta_i)) + \ell_{s}\,  \prod_{i=k}^{s}(1- p(\Delta_i))~,
\end{align*}
and notice that
\begin{align*}
E_k &\leq r_k\,\left( p(\Delta_k) + p(\Delta_{k+1}) (1- p(\Delta_k)) + \ldots +  p(\Delta_s) \prod_{i=k}^{s-1}(1- p(\Delta_i)) \right) + \ell_{s}\,  \prod_{i=k}^{s}(1- p(\Delta_i))\\
&{\mbox{(due to the fact that $r_{s} \leq r_{s-1} \leq \ldots \leq r_{k+1} \leq r_k$)}}\\
&\leq r_k\,\left( p(\Delta_k) + p(\Delta_{k+1}) (1- p(\Delta_k)) + \ldots +  p(\Delta_s) \prod_{i=k}^{s-1}(1- p(\Delta_i)) +  \prod_{i=k}^{s}(1- p(\Delta_i))\right)\\
&{\mbox{(since $\ell_s \leq 0 \leq r_k$)}}\\
& = r_k\\
&{\mbox{(since the expression in braces equals 1)~.}}
\end{align*}
Then we have, for $k\geq 2$
\begin{align}
E_{k-1} 
&= \underbrace{(1-p(\Delta_{k-1}))}_{\geq 0}\,E_k + r_{k-1}\,p(\Delta_{k-1})\label{e:backwardind}\\
&= p(\Delta_{k-1})\, \underbrace{\left(r_{k-1}-E_k\right)}_{\geq r_{k-1}-r_k \geq 0} + E_k~.\label{e:induction}
\end{align}
From (\ref{e:induction}) one can see that, viewed solely as a function of $\Delta_{k-1}$, the quantity $E_{k-1}$ can be seen as a positive constant times $p(\Delta_{k-1})$ (since $r_{k-1} - E_k \geq 0$ and $E_k$ only depends on variables $\Delta_k,\ldots, \Delta_s$) plus a constant term independent of $\Delta_{k-1}$ (again, because $E_k$ only depends on $\Delta_k,\ldots, \Delta_s$). We can now proceed by backward induction on $k = s, s-1, \ldots, 1$. For $k = s$ we have $E_s = \ell_s(1-p(\Delta_s))$ which is non-decreasing in $\Delta_s$ since so is $p(\cdot)$, and $\ell_s < 0$. Assuming by induction $E_k$ is non-decreasing in $\Delta_k, \ldots, \Delta_s$, we have from (\ref{e:induction}) that $E_{k-1}$ is non-decreasing in $\Delta_{k-1}$, thanks to the fact that $p(\Delta_{k-1})$ is monotonically increasing in $\Delta_{k-1}$, $E_{k}$ only depends on $\Delta_k\ldots, \Delta_s$, and $r_{k-1} -E_k \geq 0$. 
Moreover, $E_{k-1}$ is also non-decreasing in $\Delta_k,\ldots, \Delta_s$ since, from (\ref{e:backwardind}), $E_{k-1}$ is a positive constant (i.e., independent of $\Delta_k,\ldots,\Delta_s$) times $E_k$ plus a constant term, again independent of $\Delta_k,\ldots,\Delta_s$. Since by induction $E_k$ is non-decreasing in $\Delta_k,\ldots, \Delta_s$, so is $E_{k-1}$.

The above holds for all $k$, hence it holds in particular for $k =1$, which concludes the proof of the first part.

As for the second part, we again proceed by backward induction on $k = s, s-1, \ldots,1$.
We have $\frac{\partial E_s(\Delta_s)}{\partial \Delta_s} = -\ell_s\,\frac{\partial p (\Delta_s)}{\Delta_s} \leq z(-\ell_s) \leq z(r_s-\ell_s)$ for all $\Delta_s$. Then assume by the inductive hypothesis that 
$\frac{\partial E_k(\Delta_k,\ldots,\Delta_s)}{\partial \Delta_i} \leq z(r_i-\ell_s)$ for all $\Delta_k,\ldots,\Delta_s$, and $i = k,\ldots, s$. From (\ref{e:induction}), we can write
\begin{equation}\label{e:2z}
\frac{\partial E_{k-1}(\Delta_{k-1},\ldots, \Delta_s)}{\partial \Delta_{k-1}} = \frac{\partial p(\Delta_{k-1})}{\partial \Delta_{k-1}}(r_{k-1} - E_k) \leq z(r_{k-1}-\ell_s) \leq 2z~,
\end{equation}
the first inequality deriving from $E_k \geq  \ell_s$. On the other hand, from (\ref{e:backwardind}) we also have, for $i = k, \ldots, s$,
\[
\frac{\partial E_{k-1}(\Delta_{i},\ldots, \Delta_s)}{\partial \Delta_i} = (1-p(\Delta_{k-1}))\frac{\partial E_k(\Delta_k,\ldots, \Delta_s)}{\partial \Delta_{i}} \leq \frac{\partial E_k(\Delta_k,\ldots, \Delta_s)}{\partial \Delta_{i}} \leq z(r_i-\ell_s)~,
\]
the inequality following from the inductive hypothesis.

Again, the above holds for all $k$, hence it holds for $k =1$, which concludes the proof of the second part.

Finally, as for the third part, we first observe that, for any $k$,
\[
\frac{\partial E(\Delta_{1},\ldots, \Delta_s)}{\partial \Delta_k} = \prod_{j=1}^{k-1}(1-p(\Delta_j))\,\frac{\partial E_k(\Delta_{k},\ldots, \Delta_s)}{\partial \Delta_k} ~,
\]
and then apply the bound $\frac{\partial E_k(\Delta_{k},\ldots, \Delta_s)}{\partial \Delta_k} \leq 2z$ from (\ref{e:2z}) to obtain the claimed result.
\end{proof}

\begin{proof}[\textbf{Proof of Lemma \ref{l:bayesindep}}]
Consider the following argument.
\begin{enumerate}
\item Let $J =  \langle x_{j_{1}},x_{j_{2}},\ldots,x_{j_{k}}, \ldots, j_{k'},\ldots, x_{j_s}\rangle$, be an {\em arbitrary} sequence, and let a perturbed sequence $J' = \langle x_{j_{1}},x_{j_{2}},\ldots,x_{j_{k'}}, \ldots, x_{j_k},\ldots, x_{j_s}\rangle$ be obtained from $J$ just by swapping $x_{j_k}$ with $x_{j_{k'}}$. Moreover, suppose $p(x_{j_{k'}}) > p(x_{j_k})$. Then considering the difference 
$
\E_Y[R( J',Y)] - \E_Y[R(J,Y)]
$
and relying on the fact that rewards $r_j$ are non-decreasing, we want to show that $\E_Y[R( J',Y)] \geq \E_Y[R(J,Y)]$. It suffices to show the claim for the case where $x_{j_k}$ and $x_{j_{k'}}$ are adjacent in $J$, so that $k' = k+1$. 

Let us introduce the short-hand notation $p_i = p(x_{j_i})$, and $\Pi = \prod_{i=1}^{k-1} (1-p_i)$. Our assumption then becomes $p_{k+1} \geq p_k$.
Now, since $Y$'s components are independent, $\E_Y[R(J',Y)]$ has the form of function $E(\cdot,\ldots,\cdot)$ defined in Lemma \ref{l:monotonicity}. Then, because $k$ and $k+1$ are adjacent positions, one can easily verify that, removing common terms, the difference $\E_Y[R(J',Y)] - \E_Y[R(J,Y)]$ can be written as 
\begin{align*}
\E_Y[R(J',Y)] - \E_Y[R(J,Y)] 
&= \Pi\,\left[r_k\,(p_{k+1}-p_k) + r_{k+1}\Bigl(p_k(1-p_{k+1}) - p_{k+1}(1-p_k)\Bigl)\right]\\
&= \Pi\,(r_k-r_{k+1})(p_{k+1}-p_k) 
\end{align*}
which is non-negative, since $\Pi \geq 0$, $r_k \geq r_{k+1}$ and $p_{k+1} \geq p_k$.

As the above argument holds for an arbitrary starting sequence $J$, this shows that, for any given (unordered) set of items contained in $J$, the best way to sort them in order to maximize $\E_Y[R(J,Y)]$ is to have then in non-increasing order of their marginal probabilities $p(x_j)$.
\item Next, let $J =  \langle x_{j_{1}},x_{j_{2}},\ldots,x_{j_{k}}, \ldots, x_{j_s}\rangle$, be an {\em arbitrary} sequence, and let a perturbed sequence $J'' = \langle x_{j_{1}},x_{j_{2}},\ldots,x_{j_{k''}}, \ldots, x_{j_s}\rangle$ be obtained from $J$ just by {\em replacing} item $x_{j_k}$ by $x_{j_{k''}}$, where $p(x_{j_k''}) \geq p(x_{j_k})$. 
Again, we need to show that $\E_Y[R(\langle J'',Y)] \geq \E_Y[R(J,Y)]$. This claim immediately follows from the monotonicity property contained in Lemma \ref{l:monotonicity}, thereby showing that, for any given length $s$, the best assortment of items in $J$ is one that contains those corresponding to the $s$ largest marginal probabilities $p(x_j)$. In turn, combined with the previous item, this implies that $J^*$ has necessarily the form $J_s^* = \langle x_{j^*_{1}},x_{j^*_{2}},\ldots,x_{j^*_{s}}\rangle$, for some length $s \in \{1,\ldots,b_t\}$, where $x_{j^*_{1}},x_{j^*_{2}},\ldots,x_{j^*_{s}}$ are the items associated with the $s$ largest marginal probabilities $p(x_j)$, sorted in non-increasing order.
\item What remains is to maximize over length $s \in \{0,1,\ldots,b\}$. Notice that there is no guarantee that, viewed as a function of $s$, the quantity $ E_{Y} [R(J^*_s,Y)]$ will have a specific behavior, like unimodality. Hence, we need to try out all allowed values of $s \leq b$, including $s=0$. 
\end{enumerate}
This concludes the proof.
\end{proof}

The next lemma will be the basis for our regret analysis.
\begin{lemma}\label{l:onetimeregret}
Let us assume the independence model for outcome $Y$. Then, for given set of actions $A$, and budget $b$, let $J^*$ be the Bayes optimal sequence and $J = \langle x_{j_1},\ldots,x_{j_{s}} \rangle$ be the sequence computed by Algorithm \ref{f:2} on $A$ and $b$, with link function $\sigma$ such that $\sigma'(\Delta) \leq z$ for all $\Delta \in \R$. Further, let $\Delta_j = u^\top x_j$, and $\hDelta_j = w^\top x_j$, for all $x_j \in A$, and assume $|\Delta_j - \hDelta_j| \leq \epsilon_j$ for all $j$ such that  $x_j\in A$, where $w$ is the vector used by Algorithm \ref{f:2} to compute $J$.
Then the one-time regret
$
\E_{Y} [R(J^*,Y)] - \E_{Y} [R(J,Y)] 
$
can be bounded as follows:
\begin{align}\label{e:onetimeregret}
\E_{Y} [R(J^*,Y)] - \E_{Y} [R(J,Y)]\leq 
\begin{cases}
4z\,\sum_{i = 1}^s \epsilon_{j_i}\,\prod_{h=1}^{i-1}
(1-\sigma(\Delta_{j_h})) &{\mbox{if $J \neq \langle\rangle$}}\\
0 &{\mbox{otherwise}}~.
\end{cases}
\end{align}
\end{lemma}
\begin{proof}
Irrespective of whether $J \neq \langle\rangle$ or  $J^* \neq \langle\rangle$, we can write
\begin{eqnarray*}
\E_{Y} [R(J^*,Y)] &-& \E_{Y} [R(J,Y)]\\
&\leq&
{\wh \E_{Y}}[R(J^*,Y)] - \E_{Y}[R(J,Y)]\\
&&{\mbox{(using the first part of Lemma \ref{l:monotonicity} combined with the condition $|\Delta_j - \hDelta_j| \leq \epsilon_j$)}}\\
&\leq& 
{\wh \E_{Y}}[R(J,Y)] - \E_{Y} [R(J,Y)]\\
&&{\mbox{(since, by definition of $J$, ${\wh \E_{Y}}[R(J^*,Y)] \leq {\wh \E_{Y}}[R(J,Y)]$.}}\\
&&{\mbox{Notice that this implies that in the case where our algorithm happens to play $J = \langle\rangle$}} \\
&&{\mbox{the regret is $\leq 0$).}}\\
&=&
E(\hDelta_{j_1}+\epsilon_{j_1},\ldots,\hDelta_{j_s}+\epsilon_{j_s})-
E(\Delta_{j_1},\ldots, \Delta_{j_s})\\
&&{\mbox{(where $E(\cdot)$ is defined in (\ref{e:ehat}))}}\\
&\leq&
E(\Delta_{j_1}+2\epsilon_{j_1},\ldots,\Delta_{j_s}+2\epsilon_{j_s}) 
-
E(\Delta_{j_1},\ldots,\Delta_{j_s})\\
&&{\mbox{(using again the first part of Lemma \ref{l:monotonicity} together with $|\Delta_j - \hDelta_j| \leq \epsilon_j$)}}~.
\end{eqnarray*}
%

%

Now, by the mean-value theorem, we can write
\[
E(\Delta_{j_1}+2\epsilon_{j_1},\ldots,\Delta_{j_s}+2\epsilon_{j_s}) 
-
E(\Delta_{j_1},\ldots,\Delta_{j_s}) 
=
2\,\sum_{i = 1}^s \frac{\partial E(\Delta_{j_1},\ldots, \Delta_{j_s})}{\partial \Delta_{j_i}}{\big|}_{\Delta_{j_1} = \xi_{j_s},\ldots,\Delta_{j_s} = \xi_{j_s} }\, \epsilon_{j_i}~,
\]
where $\xi_{j_i} \in (\Delta_{j_i},\Delta_{j_i}+2\epsilon_{j_i})$, for $i \in [s]$.
The third part of Lemma \ref{l:monotonicity} then allows us to write
\begin{eqnarray*}
\frac{\partial E(\xi_{j_1},\ldots, \xi_{j_s})}{\partial \Delta_{j_i}}
&\leq&
2z\, (1-\sigma(\xi_{j_1}))\ldots (1-\sigma(\xi_{j_{i-1}})) \\
&\leq&
2z\, (1-\sigma(\Delta_{j_1}))\ldots (1-\sigma(\Delta_{j_{i-1}}))~,
\end{eqnarray*}
the second inequality deriving from the monotonicity of $\sigma(\cdot)$ and the fact that $\xi_{j_i} \in (\Delta_{j_i},\Delta_{j_i}+2\epsilon_{j_i})$.
Replacing back, and summing over $i$ yields the claimed bound.
\end{proof}

\begin{lemma}\label{l:biasvariance}
Consider any item $x_{j_i} \in A$, and the random variable $s_{j_i} \in \{-1,0,1\}$ whose value is given in the algorithm's pseudocode. Also, assume $x_{j_i}$ occurs in the $i$-th position of sequence $J = \langle x_{j_1}, x_{j_2}, \ldots, x_{j_s}\rangle$. Let $c_{\sigma}$ and $c_{\sigma'}$ be two positive constants such that, for all $\Delta \in [-D,D]$ we have $|L'(\Delta)| \leq c_{\sigma}$ and $L''(\Delta) \geq c_{\sigma'}$. Set $\Delta_{j_i} = u^\top x_{j_i}$.
Then, for any ${\hat \Delta'_{j_i}} \in \R$ we have
\[
0 \leq \var [L(s_{j_i}{\hat \Delta_{j_i}}) - L(s_{j_i}\Delta_{j_i})\,|\, J] 
\leq \frac{2(c_{\sigma})^2}{c_{\sigma'}}\,\E [L(s_{j_i}{\hat \Delta_{j_i}}) - L(s_{j_i}\Delta_{j_i})\,|\, J]~.
\]
\end{lemma}
\begin{proof}
Let us introduce the shorthands 
\[
\Delta_{j} = u_j^\top x,\qquad p_{j_i} = \sigma(\Delta_{j_i}),\qquad \Pi_{i-1} = (1-\sigma(\Delta_{j_1})) \ldots (1-\sigma(\Delta_{j_{i-1}}))~.
\]
We can write
\begin{eqnarray*}
\PP(s_{j_i} = 1\,|\, J) &=&  \Pi_{i-1}\, p_{j_i}\\
\PP(s_{j_i} = -1\,|\, J) &=& \Pi_{i-1}\, (1-p_{j_i})\\
\PP(s_{j_i} = 0\,|\, J) &=& 1- \PP(s_{j_i} = 1\,|\, J) - \PP(s_{j_i} = -1\,|\, J)~.
\end{eqnarray*}
Hence, for all ${\hat \Delta_{j_i}} \in \R$ we have
\begin{eqnarray*}
\E [L(s_{j_i}{\hat \Delta_{j_i}}) &-& L(s_{j_i}\Delta_{j_i})\,|\, J] \\
&=& \Pi_{i-1}\left(\, p_{j_i} \left(L({\hat \Delta_{j_i}}) - L(\Delta_{j_i})\right)
+  (1-p_{j_i}) \left(L(-{\hat \Delta_{j_i}}) - L(-\Delta_{j_i})\right) \right)\\
&\geq&
\Pi_{i-1}\Bigl(\, p_{j_i} 
\left(L'(\Delta_{j_i})({\hat \Delta_{j_i}}- \Delta_{j_i}) + \frac{c_{\sigma'}}{2}\,({\hat \Delta_{j_i}}- \Delta_{j_i})^2\right)\\
&&\qquad\qquad+  (1-p_{j_i}) \left( L'(-\Delta_{j_i})(\Delta_{j_i}- {\hat \Delta_{j_i}}) + \frac{c_{\sigma'}}{2}\,({\hat \Delta_{j_i}}- \Delta_{j_i})^2\right) \Bigl)\\
&&{\mbox{(using $L''(\Delta_{j_i}) \geq c_{\sigma'}$)}}\\
&=&\Pi_{i-1}\frac{c_{\sigma'}}{2}\,({\hat \Delta_{j_i}}- \Delta_{j_i})^2\\
&&{\mbox{(since $p_{j_i} = L'(-\Delta_{j_i})$ and $1-p_{j_i} = L'(\Delta_{j_i})$}}~.
\end{eqnarray*}
Moreover,
\begin{eqnarray*}
\var [L(s_{j_i}{\hat \Delta_{j_i}}) - L(s_{j_i}\Delta_{j_i})\,|\, J] 
&\leq&
\E[(L(s_{j_i}{\hat \Delta_{j_i}}) - L(s_{j_i}\Delta_{j_i}))^2\,|\, J]\\
&\leq& \Pi_{i-1} (c_{\sigma})^2 ({\hat \Delta_{j_i}}-\Delta_{j_i})^2\\
&&{\mbox{(using $|L'(\Delta_{j_i})| \leq c_{\sigma}$)}}~.
\end{eqnarray*}
Piecing together gives the claimed bound.
\end{proof}

The next lemma helps us define the upper confidence parameters $\epsilon_{j,t}$. To this effect, for $t \in [T]$, let $d_{t}(u,w)$ be the Mahalanobis distance  between vectors $u$ and $w$ as
\[
d_{c_t}(u,w) = (u - w)^\top M_{c_t}  (u - w)~,
\]
where $M_{c_t}$ is the matrix maintained by Algorithm \ref{f:2} at the $c_t$-th update.
In order to quantify $\epsilon_{j}$ in Lemma \ref{l:onetimeregret}, we introduce a suitable surrogate loss function $L(\cdot)$ that determines the dynamics of the algorithm (i.e., the proposed update rule being an online Newton step w.r.t. to this loss function), along with its convergence guarantees. In the proof of this lemma (see the appendix) we set $L(\Delta) =\log(1+e^{-\Delta})$. Notice that $\sigma(\Delta) = -L'(-\Delta)$.
The lemma follows indeed from somewhat standard arguments, and relies on the exp-concavity of $L(\cdot)$. 
\begin{lemma}\label{l:mahalanobis}
Assume there exists $D >0$ such that $u^\top x_j \in [-D,D]$ for all $x_j \in A$.
Let $c_{\sigma}$ and $c_{\sigma'}$ be two positive constants such that, for all $\Delta \in [-D,D]$ we have 
$0 < 1-c_{\sigma} \leq \sigma(\Delta) \leq c_{\sigma} < 1$ and $\sigma'(\Delta) \geq c_{\sigma'}$.
Then with probability at least $1-\delta$, with $\delta < 1/e$, we have
\begin{align*}
&d_{c_t-1}(u,w'_{c_t})\\ &\leq 
bD^2 
+ \left(\frac{c_{\sigma}}{c_{\sigma'}}\right)^2 d\log \left(1+ \frac{2}{b}\Bigl(\frac{t\,c_{\sigma}}{1-c_{\sigma}} + 4\log\frac{2(t+1)}{\delta}\Bigl) \right)
 + \left(12\left(\frac{c_{\sigma}}{c_{\sigma'}}\right)^2 + \frac{36(1+D)}{c_{\sigma'}} \right) \log \frac{2b(t+4)}{\delta}~
\end{align*}
uniformly over $c_t \in [bT]$, where $b_t \leq b$ for all $t \in [T]$.
\end{lemma}
\begin{proof}
Given items $A$, the update rules $w'_{c_t+j-1} \rightarrow w_{c_t+j} \rightarrow w'_{c_t+j}$ combined with the lower bound $L''(\Delta) \geq c_{\sigma'}$ allows us to write for all $t$ (adapted from, e.g.,~\cite{DBLP:journals/ml/HazanAK07,DBLP:journals/corr/abs-1207-0166})
\begin{align}\label{e:hazan+}
&d_{c_{t}-1}(u,w'_{c_t}) \notag\\
&\leq 
b D^2 + \left(\frac{1}{c_{\sigma'}}\right)^2\cdot \sum_{k=1}^{t-1} \sum_{j=1}^{{\hat s_k}} \nabla_{j,k}^\top M_{c_k+j-1}^{-1}\nabla_{j,k}  - \frac{2}{c_{\sigma'}} \sum_{k=1}^{t-1}\sum_{j=1}^{{\hat s_k}} \Bigl(L(s_{j,k}x_j^\top w'_{c_k+j-1} ) - L(s_{j,k} u^\top x_j ) \Bigl)~,
\end{align}
where $c_k = {\hat s_1} + {\hat s_2} + \ldots + {\hat s_{k-1}}$.

In particular, notice that the step $w_{c_t+j} \rightarrow w'_{c_t+j}$ is a projection of $w_{c_t+j}$ onto the convex set $\{w\in \R^d\,:\, -D \leq w^\top x_{j} \leq D\}$ w.r.t. Mahalanobis distance $d_{c_t+j-1}(\cdot,\cdot)$. This projection can be computed in closed form as follows:
$$
w'_{c_t+j-1} = 
\begin{cases}
w_{c_t+j-1} &{\mbox{if $|w_{c_t+j-1}^\top x_j| \leq D$}}\\
w_{c_t+j-1} - \frac{w_{c_t+j-1}^\top x_j-D}{x_j^\top M_{c_t+j-2}^{-1}x_j} M^{-1}_{c_t+j-2}x_j &{\mbox{if $w_{c_t+j-1}^\top x_j > D$}}\\
w_{c_t+j-1} - \frac{w_{c_t+j-1}^\top x_j+D}{x_j^\top M_{c_t+j-2}^{-1}x_j} M^{-1}_{c_t+j-2}x_j &{\mbox{if $w_{c_t+j-1}^\top x_j < -D$~.}}
\end{cases}
$$
Further, we lower bound with high probability $\sum_{k=1}^{t-1}\sum_{j=1}^{{\hat s_k}} \Bigl(L(s_{j,k}x_j^\top w'_{c_k+j-1} ) - L(s_{j,k} u^\top x_j ) \Bigl)$ using the fact that the conditional expectation of the loss difference $L(s_{j,k}x_j^\top w'_{c_k+j-1} ) - L(s_{j,k} u^\top x_j )$ is non-negative (Lemma \ref{l:biasvariance}).\footnote
{
Here, Lemma \ref{l:biasvariance} is applied with expectations conditioned on past history.
} 
The same lemma also allows for fast rates of convergence, so that we can apply any Freedman-like inequality (see, e.g., Lemma 3 in \cite{DBLP:conf/nips/KakadeT08}) for bounded martingale difference sequences to conclude that
\[
\sum_{k=1}^{t-1}\sum_{j=1}^{{\hat s_k}} \Bigl(L(s_{j,k}x_j^\top w'_{c_k+j-1} ) - L(s_{j,k} u^\top x_j ) \Bigl)
\geq 
- \left(\frac{6(c_{\sigma})^2}{c_{\sigma'}} + 18L(-D) \right) \log \frac{b(t+4)}{\delta}
\]
with $b \geq b_t$ for all $t$,
holds with probability $\geq 1-\delta/(bt(t+1))$, the boundedness of the difference sequence following from the fact that $|u^\top x_j| \leq D$ holds by assumption, and $|x_j^\top w'_{c_k+j-1}| \leq D$ holds by the projection steps $w_{c_k+j-1} \rightarrow w'_{c_k+j-1}$.
We then upper bound $L(-D)$ by $1+D$ and exploit a known upper bound:
\begin{align}
\sum_{k=1}^{t-1} \sum_{j=1}^{{\hat s_k}} \nabla_{j,k}^\top M_{c_k+j-1}^{-1}\nabla_{j,k}
&=
\sum_{k=1}^{t-1} \sum_{j=1}^{{\hat s_k}}  
\sigma^2(-s_{j,k} x_j^\top w'_{c_k+j-1}) |s_{j,k}| \left(x_{j}^\top M_{c_k+j-1}^{-1} x_j\right)\notag\\ 
&\leq
(c_{\sigma})^2\,\sum_{k=1}^{t-1} \sum_{j=1}^{{\hat s_k}}  |s_{j,k}| \left(x_{j}^\top M_{c_k+j-1}^{-1} x_j\right)\notag\\
&{\mbox{(from the fact that $L'(\Delta) \leq c_{\sigma}$ for all $\Delta \in [-D,D]$, and $|x_j^\top w'_{c_k+j-1}| \leq D$)}}\notag\\
&\leq
(c_{\sigma})^2 d\log \left(1+ \frac{1}{b}\,\sum_{k=1}^{t-1} \sum_{j=1}^{{\hat s_k}} |s_{j,k}| \right)\label{e:standardineq}\\
&{\mbox{(from a standard inequality, e.g., \cite{DBLP:journals/corr/abs-1301-6677,DBLP:journals/siamcomp/Cesa-BianchiCG05})~.}}\notag
\end{align}
Since $|s_{j,k}|$ is a Bernoulli random variable which is 1 (that is, the corresponding component of outcome vector $Y_k$ is observed) with (conditional) probability 
$\Pi_{j-1,k} = \prod_{i = 1}^{j-1} (1-\sigma(\Delta_{i,k}))~,$
where
\[
\Delta_{i,k} = u^\top x_{i}~, ~~~~~i = 1, \ldots, {\wh s_k}~,
\]
we can apply again the aforementioned
Freedman-like inequality from \cite{DBLP:conf/nips/KakadeT08} to conclude that
\[
\Pr\left(\exists t\,:\,\sum_{k=1}^{t-1} \sum_{j=1}^{{\hat s_k}} |s_{j,k}| \leq 2\sum_{k=1}^{t-1} \sum_{j=1}^{{\hat s_k}} \Pi_{j-1,k} + 4\log \frac{t(t+1)}{\delta} \right) \geq 1-\delta~.
\]
In turn, since $\Delta_{i,k} \in [-D,D]$, we have $1-\sigma(\Delta_{i,k}) \leq c_{\sigma}$ for all $i$ and $k$, so that $\sum_{j=1}^{\wh s_t}\Pi_{j-1,k} \leq \sum_{j=1}^{\infty}\left(c_{\sigma}\right)^i = \frac{c_{\sigma}}{1-c_{\sigma}}$. After some overapproximations, the above implies
\[
\Pr\left(\exists t\,:\,\sum_{k=1}^{t-1} \sum_{j=1}^{{\hat s_k}} |s_{j,k}| \leq 2(t-1)\,\frac{c_{\sigma}}{1-c_{\sigma}}  + 8\log \frac{t+1}{\delta} \right) \geq 1-\delta~.
\]
We plug it back into (\ref{e:standardineq}), then
%
%
back into (\ref{e:hazan+}) and replace $\delta$ by $\delta/2$ to obtain the claimed result.
\end{proof}

\begin{lemma}\label{l:adjust}
Let $M$ be a $d\times d$ positive definite matrix whose minimal eigenvalue is $\geq b$, for some $b \in \{1,2,\ldots,\}$, 
and $x_1, x_2, \ldots, x_b \in \{x\in \R^d\,:\, ||x||\leq 1\}$. 
Then
\[
\sum_{j = 1}^b 
x_j^\top M^{-1} x_j \leq e\,\sum_{j = 1}^b 
x_j^\top M_{j}^{-1} x_j~,
\]
where $M_j = M + x_1x_1^\top + \ldots x_j x_j^\top$, and $e$ is the base of natural logarithms.
\begin{proof}
Consider the quantity $x^\top M_{j}^{-1} x$, with $M_0 = M$. We first prove that, for any $x\in \R^d$,
\begin{equation}\label{e:rec}
x^\top M^{-1} x \leq \left(1+\frac{1}{b}\right)^{j}\,x^\top M_{j}^{-1} x
\end{equation}
holds for all $j \in [b]$.

By the Sherman-Morrison formula for matrix inversion we have, for an arbitrary $x \in \R^d$, and $j \geq 1$,
\begin{align*}
x^\top M_{j}^{-1} x 
&= x^\top (M_{j-1} + x_{j}x_{j}^\top)^{-1} x\\ 
&= x^\top M_{j-1}^{-1}x  - 
\frac{(x^\top M_{j-1}^{-1}x_{j})^2}{1+x_{j}^\top M_{j-1}^{-1}x_{j}}\\
&\geq 
x^\top M_{j-1}^{-1}x  - 
\frac{(x^\top M_{j-1}^{-1}x)(x_{j}^\top M_{j-1}^{-1}x_{j})}{1+x_{j}^\top M_{j-1}^{-1}x_{j}}\\
&{\mbox{(from the Cauchy-Schwarz inequality)}}
\end{align*}
so that
\[
x^\top M_{j-1}^{-1}x \leq x^\top M_{j}^{-1} x + \frac{(x^\top M_{j-1}^{-1}x)(x_{j}^\top M_{j-1}^{-1}x_{j})}{1+x_{j}^\top M_{j-1}^{-1}x_{j}}~.
\]
Hence, rearranging terms, we can write
\[
x^\top M_{j-1}^{-1}x \leq x^\top M_{j}^{-1} x (1+x_{j}^\top M_{j-1}^{-1}x_{j}) \leq x^\top M_{j}^{-1} x \left(1+\frac{1}{b}\right),
\]
the second inequality deriving from the assumption $||x_j|| \leq 1$ and the fact that since the smallest eigenvalue of $M$ is at least $b$, so is the smallest eigenvalue of $M_{j-1} \geq M$.
Unwrapping this recurrence over $j$ gives (\ref{e:rec}).

From  (\ref{e:rec}), since
$(1+1/b)^{j} \leq e$ when $j \leq b$, we have
\[
x^\top M^{-1}x \leq  e\,x^\top M_{j}^{-1} x~.
\]
Since this holds for a generic $x$, we instantiate in turn $x$ to $x_1, x_1,\ldots, x_b$, 
and sum over $j \in [b]$. This yields
\[
\sum_{j = 1}^b 
x_j^\top M^{-1} x_j \leq e\,\sum_{j = 1}^b 
x_j^\top M_{j}^{-1} x_j~,
\]
as claimed.
\end{proof}
\end{lemma}

\begin{proof}[\textbf{Proof of Theorem \ref{t:independent}}]
Consider matrix $M_{c_t-1}$ in Lemma \ref{l:mahalanobis}. If $J_r = \langle x_{\wh j_{r,1}}, \ldots, x_{\wh j_{r,{\wh s_r}}} \rangle$, for $r=1,\ldots, t-1$, we can write
\[
M_{c_t-1} = bI + \sum_{r=1}^{t-1}\sum_{j=1}^{\wh s_r} |s_{j,r}|\,x_{\wh j_{r},j} x_{\wh j_{r},j}^\top~,
\]
where $|s_{j,r}|$ is a Bernoulli random variable which is 1 (that is, the corresponding component of outcome vector $Y_r$ is observed) with probability 
$\Pi_{j-1,r} = \prod_{i = 1}^{j-1} (1-\sigma(\Delta_{i,r}))~,$
where
\[
\Delta_{i,r} = u^\top x_{\wh j_r,i}~, ~~~~~i = 1, \ldots, {\wh s_r}~.
\]
Let 
\[
{\bar M_{c_t-1}} = bI + \sum_{r=1}^{t-1}\sum_{j=1}^{\wh s_r} \Pi_{j-1,r}\,x_{\wh j_{r},j} x_{\wh j_{r},j}^\top~,
\]
and consider the matrix martingale difference sequence
\[
|s_{j,r}|\,x_{\wh j_{r},j}x_{\wh j_{r},j}^\top - \Pi_{j-1,r}\,x_{\wh j_{r},j}x_{\wh j_{r},j}^\top~,\qquad r= 1,\ldots,t-1,\,j=1,\ldots,{\wh s_r}~.
\]
By a standard Freedman-style matrix martingale inequality (e.g., \cite{tropp11}) adapted to our scenario we have, for positive constants $\theta$ and $\theta'$,
\begin{equation}\label{e:tropp}
\Prob{\exists t\,:\, \lambda_{\max}\left(M_{c_t-1} - {\bar M_{c_t-1}}\right)  \geq \theta,~ ||{\bar M_{c_t-1}}|| \leq \theta'} 
\leq d\,\exp\left(\frac{-\theta^2/2}{\theta'+\theta/3}\right)~,
\end{equation}
where $\lambda_{\max}(\cdot)$ denotes the algebraically largest eigenvalue of the matrix at argument, and $||\cdot||$ denotes the spectral norm.

We now proceed according to a standard stratification argument (e.g., \cite{DBLP:journals/tit/Cesa-BianchiG08}) Setting $A(x,\delta) = 2\log\frac{xd}{\delta}$ and $f(A,r) = 2A +\sqrt{Ar}$, we can write
\begin{align}
&\Prob{\exists t\,:\, \lambda_{\max}\left( M_{c_t-1} - {\bar M_{c_t-1}}\right) \geq  f(A(||{\bar M_{c_t-1}}||,\delta),||{\bar M_{c_t-1}}||) }\notag\\
&\leq
\sum_{r=0}^{\infty} \Prob{\exists t\,:\, \lambda_{\max}\left( M_{c_t-1} - {\bar M_{c_t-1}}\right) \geq  f(A(||{\bar M_{c_t-1}}||,\delta),||{\bar M_{c_t-1}}||),~~ 2^r-1 \leq ||{\bar M_{c_t-1}}|| \leq 2^{r+1} }\notag\\
&\leq
\sum_{r=0}^{\infty} \Prob{\exists t\,:\, \lambda_{\max}\left( M_{c_t-1} - {\bar M_{c_t-1}}\right) \geq  f(A(2^{r+1} ,\delta),2^{r+1}),~~||{\bar M_{c_t-1}}|| \leq 2^{r+1} }\notag\\
&\leq
\sum_{r=0}^{\infty}  d\,\exp\left(\frac{-f^2(A(2^{r+1} ,\delta),2^{r+1})/2}{2^{r+1}+f(A(2^{r+1} ,\delta),2^{r+1})/3}\right)~,\notag
\end{align}
the last inequality deriving from (\ref{e:tropp}).

Since $f(A, r)$ satisfies $f^2(A,r) \geq Ar + A + 2/3f(A, r)A$, the exponent in the last exponential is at least $A(2^{r+1}, \delta)/2$, implying
\[
\sum_{r=0}^{\infty} \exp\left(- A(2^{r+1}, \delta)/2\right) = \sum_{r=0}^{\infty} \frac{\delta}{d\,2^{r+1}} = \delta/d~,
\]
which in turn implies
\[
\Prob{\exists t\,:\, \lambda_{\max}\left( M_{c_t-1} - {\bar M_{c_t-1}}\right) \geq  f(A(||{\bar M_{c_t-1}}||,\delta),||{\bar M_{c_t-1}}||) } \leq \delta~.
\]
Plugging back the definitions of $f(A,r)$ and $A(x,\delta)$, noticing that, $||{\bar M_{c_t-1}}|| = \lambda_{\max}({\bar M_{c_t-1}}) \leq b(t+1)$ (due to the fact that $||{\bar M_{c_t-1}}||$ is positive definite and $||x_{\wh j_t,j}|| \leq 1$), and overapproximating gives
\[
\Prob{\exists t\,:\, \lambda_{\max}\left( M_{c_t-1} - {\bar M_{c_t-1}}\right) \geq 4\log\frac{bd(t+1)}{\delta} +\sqrt{2\,\lambda_{\max}({\bar M_{c_t-1}})\,\log \frac{bd(t+1)}{\delta}  }  } \leq \delta~.
\]
Further, we use $\sqrt{ab} \leq a/2+b/2$ with $a=\lambda_{\max}({\bar M_{c_t-1}})$ and $b = 2\log \frac{bd(t+1)}{\delta}$~. 
Rearranging gives
\[
\Prob{\exists t\,:\, \frac{1}{2}\,\lambda_{\max}\left( {\bar M_{c_t-1}} \right) -  
\lambda_{\max}({\bar M_{c_t-1}} - M_{c_t-1}) \leq
-5\,\log\frac{bd(t+1)}{\delta}} \leq \delta
\]
or
\[
\Prob{\forall t\,\, \frac{1}{2}\,\lambda_{\max}\left( {\bar M_{c_t-1}} \right) -  
\lambda_{\max}({\bar M_{c_t-1}} - M_{c_t-1}) \geq
-5\,\log\frac{bd(t+1)}{\delta}} 
\geq 1-\delta~.
\]
Now, observing that
\begin{align*}
\lambda_{\max}\left( {\bar M_{c_t-1}} \right) -  
\lambda_{\max}(2{\bar M_{c_t-1}} - 2M_{c_t-1}) 
& \leq 
\lambda_{\max}\left(2 M_{c_t-1} - {\bar M_{c_t-1}}\right)
\end{align*}
the above implies
\[
\Prob{\forall t\,\, \lambda_{\max}\left( M_{c_t-1} -  \frac{1}{2}{\bar M_{c_t-1}}\right) \geq
-5\,\log\frac{bd(t+1)}{\delta}} 
\geq 1-\delta~,
\]
which can be rewritten as
\[
\Prob{\forall t\,\,\forall v \in \R^d\,:\, \frac{v^\top\left( M_{c_t-1} -  \frac{1}{2}{\bar M_{c_t-1}}\right)v}{v^\top v} \geq
-5\,\log\frac{bd(t+1)}{\delta}} 
\geq 1-\delta~.
\]

If we define 
\[
{\bar d}_{c_t-1}(u,w) = (u-w)^\top {\bar M_{c_t-1}} (u-w)
\]
the above inequality allows us to conclude that \[
d_{c_t-1}(u,w) \geq \frac{1}{2}\,{\bar d}_{c_t-1}(u,w) - 20D^2\log\frac{bd(t+1)}{\delta}
\]
holds with probability at least $1-\delta$, uniformly over all $u,w \in \R^d$ such that $||u-w|| \leq 2D$ and all rounds $t$. Hence, combining with Lemma \ref{l:mahalanobis}, and upper bounding $t$ by $T$,
\[
{\bar d}_{c_t-1}(u,w'_{c_t}) \leq \alpha(b,d,T,2\delta)
\]
where
\begin{align*}
\alpha(b,d,T,2\delta) = 2bD^2 
&+ 
\left(\frac{c_{\sigma}}{c_{\sigma'}}\right)^2 d\log \left(1+ \frac{2}{b}\Bigl(\frac{T\,c_{\sigma}}{1-c_{\sigma}} + 4\log\frac{2(T+1)}{\delta}\Bigl) \right)\\
&+ 2\left(12\left(\frac{c_{\sigma}}{c_{\sigma'}}\right)^2 + \frac{36(1+D)}{c_{\sigma'}} \right) \log \frac{b(T+4)}{\delta} 
+  20D^2\log\frac{bd(T+1)}{\delta}
\end{align*}
with probability at least $1-2\delta$.

Then Cauchy-Schwarz inequality allows us to write, for all $x \in \R^d$,
\[
(u^\top x - x^\top w'_{c_t})^2 \leq x^\top {\bar M}_{c_t-1}^{-1}x\, {\bar d}_{c_t-1}(u,w'_{c_t})
\leq \left(x^\top {\bar M}_{c_t-1}^{-1}x\right)\,\alpha(b,d,T,2\delta) ~.
\]

We are therefore in a position to apply Lemma \ref{l:onetimeregret} with $J$ therein set to $J_t = \langle x_{\wh j_{t,1}}, \ldots, x_{\wh j_{t,{\wh s_t}}} \rangle $ and $\epsilon_j$ set to $\epsilon_{\wh j_{t},j} = \sqrt{\left(x_{\wh j_t,j}^\top {\bar M}_{c_t-1}^{-1}x_{{\wh j_t,j}}\right)\,\alpha(b,d,T,2\delta)}$,~~ for $j = 1, \ldots, {\wh s_t}$. Thus we can write
\begin{equation}\label{e:regretbound}
\sum_{t=1}^T \E_{Y_t} [R(J^*_t,Y_t)] - \E_{Y_t} [R(J_t,Y_t)] 
\leq 
4z\sqrt{\alpha(b,d,T,2\delta)}\,\sum_{t=1}^T \sum_{j = 1}^{{\wh s_t}} \sqrt{\left(x_{\wh j_t,j}^\top {\bar M}_{c_t-1}^{-1}x_{{\wh j_t,j}}\right)} \,\Pi_{j-1,t}~.
\end{equation}

Now, for each round $t$, consider the quantity
\[
\sum_{j = 1}^{{\wh s_t}} \left(x_{\wh j_t,j}^\top {\bar M}_{c_t-1}^{-1}x_{{\wh j_t,j}}\right) \,\Pi_{j-1,t}
\]
Noticing that ${\bar M}_0 = bI$, we invoke Lemma \ref{l:adjust} with $x_j$ therein set to $x_{\wh j_t,j}\sqrt{\Pi_{j-1,t}}$ and write
\begin{equation}\label{e:useoflemma}
\sum_{j = 1}^{{\wh s_t}} \left(x_{\wh j_t,j}^\top {\bar M}_{c_t-1}^{-1}x_{{\wh j_t,j}}\right) \,\Pi_{j-1,t} 
\leq 
e\,\sum_{j = 1}^{{\wh s_t}} \left(x_{\wh j_t,j}^\top {\bar M}_{c_t-1+j}^{-1}x_{{\wh j_t,j}}\right) \,\Pi_{j-1,t}
\end{equation}
where
\[
{\bar M}_{c_t-1+j} = {\bar M}_{c_t-1} + \sum_{i=1}^{j}x_{{\wh j_t,i}} x_{{\wh j_t,i}}^\top  \Pi_{i-1,t}~,
\]
with $\Pi_{0,t} = 1$. Thus, for each $t$,
\begin{align*}
\sum_{j = 1}^{{\wh s_t}} \sqrt{\left(x_{\wh j_t,j}^\top {\bar M}_{c_t-1}^{-1}x_{{\wh j_t,j}}\right)} \,\Pi_{j-1,t} 
&= 
\sum_{j = 1}^{{\wh s_t}} \sqrt{\left(x_{\wh j_t,j}^\top {\bar M}_{c_t-1}^{-1}x_{{\wh j_t,j}}\right)\Pi_{j-1,t}} \,\sqrt{\Pi_{j-1,t}}\\
&\leq
\sqrt{\sum_{j = 1}^{{\wh s_t}} \left(x_{\wh j_t,j}^\top {\bar M}_{c_t-1}^{-1}x_{{\wh j_t,j}}\right)\Pi_{j-1,t}} \,\sqrt{\sum_{j = 1}^{{\wh s_t}} \Pi_{j-1,t}}\\
&{\mbox{(from the Cauchy-Schwarz inequality)}}\\
&\leq
\sqrt{\frac{e\,c_{\sigma}}{1-c_{\sigma}}\,\sum_{j = 1}^{{\wh s_t}} \left(x_{\wh j_t,j}^\top {\bar M}_{c_t-1+j}^{-1}x_{{\wh j_t,j}}\right)\Pi_{j-1,t}}\\
&{\mbox{(from (\ref{e:useoflemma}), along with $\Pi_{j-1,t} \leq 1$ and 
$\sum_{j=1}^{\wh s_t}\Pi_{j-1,t} \leq \frac{c_{\sigma}}{1-c_{\sigma}}$,}}\\
&{\mbox{as argued within the proof of Lemma \ref{l:mahalanobis})~.}}
\end{align*}

Getting back to (\ref{e:regretbound}), 
combining with the last inequality we have
\begin{align*}
\sum_{t=1}^T \E_{Y_t} [R(J^*_t,Y_t)] &- \E_{Y_t} [R(J_t,Y_t)] \\
&\leq 
4z\sqrt{\alpha(b,d,T,2\delta)}\,\sum_{t=1}^T \sqrt{\frac{e\,c_{\sigma}}{1-c_{\sigma}}\,\sum_{j = 1}^{{\wh s_t}} \left(x_{\wh j_t,j}^\top {\bar M}_{c_t-1+j}^{-1}x_{{\wh j_t,j}}\right)\Pi_{j-1,t}}\\
&\leq 
4z\sqrt{\frac{e\,c_{\sigma}}{1-c_{\sigma}} \alpha(b,d,T,\delta)T}\,\sum_{t=1}^T \sum_{j = 1}^{{\wh s_t}} \left(x_{\wh j_t,j}^\top {\bar M}_{c_t-1+j}^{-1}x_{{\wh j_t,j}}\right)\Pi_{j-1,t}~,\\
&{\mbox{(again from the Cauchy-Schwarz inequality)}}\\
&\leq
4z\sqrt{\frac{e\,c_{\sigma}}{1-c_{\sigma}} \alpha(b,d,T,\delta)T}\,d\,\log \left(1 + \frac{1}{b}\sum_{t=1}^T \sum_{j = 1}^{{\wh s_t}} \Pi_{j-1,t}   \right)\\
&{\mbox{(from a standard inequality, e.g., \cite{DBLP:journals/corr/abs-1301-6677,DBLP:journals/siamcomp/Cesa-BianchiCG05,DBLP:conf/nips/Abbasi-YadkoriPS11},~along with $||x_{{\wh j_t,j}}||  \leq 1$ and $M_0 = bI$)}}\\
&\leq
4z\sqrt{\frac{e\,c_{\sigma}}{1-c_{\sigma}}\alpha(b,d,T,\delta)T}\,d\,\log(1+T)\\
&{\mbox{(since $\Pi_{j-1,t} \leq 1$ and ${\wh s_t} \leq b$)}}~.
\end{align*}
Since the above holds with probability $\geq 1-2\delta$, we replace $\delta$ by $\delta/2$ in $\alpha(b,d,T,2\delta)$ so as to obtain the claimed result.

\end{proof}

\section{Algorithm for the Case of Dependent Outcomes}\label{as:dependent}
For completeness, we give in Algorithm \ref{a:dependent} the pseudocode of the greedy algorithm used in our experiments. 
All in all, the algorithm performs the same updates as Algorithm \ref{f:1}, but applied to the coverage difference vectors ${\bar c}(x_{j_k} \mid x_{j_1}, \ldots, x_{j_{k-1}})$ instead of the original feature vectors $x_{j_k}$. Moreover, Algorithm \ref{a:dependent} replaces the computation of $J_t$ by mimicking {\sc Greedy}, described in Section \ref{ss:dependent}.

In the pseudocode of Algorithm \ref{a:dependent} we define
\begin{align*}
\alpha(b,d',T,\delta) = 2bD^2 
&+ \left(\frac{c_{\sigma}}{c_{\sigma'}}\right)^2 d'\log \left(1+ \frac{2}{b}\Bigl(\frac{T\,c_{\sigma}}{1-c_{\sigma}} + 4\log\frac{4(T+1)}{\delta}\Bigl) \right)\\
&+ 2\left(12\left(\frac{c_{\sigma}}{c_{\sigma'}}\right)^2 + \frac{36(1+D)}{c_{\sigma'}} \right) \log \frac{2b(T+4)}{\delta}
+  20D^2\log\frac{2bd'(T+1)}{\delta}~.
\end{align*}

\begin{algorithm*}[t!]
{\bf Input:} Confidence level $\delta \in [0,1]$,  width parameter $D > 0$, maximal budget parameter $b>0$;\\
{\bf Init:} $M_{0} = bI \in \R^{d'\times d'}$,\ $w_{1} = 0\in \R^{d'}$, $c_1 = 1$\\[0mm]
{\bf For } $t=1, 2,\ldots, T$ :
\begin{enumerate}
\vspace{-0.05in}
\item Get:
\vspace{-0.05in}
\begin{itemize}
    \item Set of actions $A_t = \{x_{1,t},\ldots, x_{|A_t|,t}\} \subseteq \{  x \in \R^{d'}\,:\,||x|| \leq 1\}$~,
    \vspace{-0.05in}
    \item budget $b_t \leq b$~;
\end{itemize}
\vspace{-0.05in}
\item Compute $J_t$~:
\begin{itemize}
\vspace{-0.05in}
\item {\bf For } $k=1, \ldots, \min\{b_t,|A_t|\}$ :
        \vspace{-0.05in}
        \[
        x_{\wh j_{t,k}} = \argmax_{x \in A_t\setminus\{x_{\wh j_{t,1}}, \ldots, x_{\wh j_{t,k-1}}\}} \ \sigma\Bigl( {\bar c}(x \mid x_{\wh j_{t,1}}, \ldots, x_{\wh j_{t,k-1}})^\top w_{c_t} + \epsilon_t(x \mid x_{\wh j_{t,1}}, \ldots, x_{\wh j_{t,k-1}}) \Bigl)~,
        \]
        where $\epsilon_t^2(x \mid x_{\wh j_{t,1}}, \ldots, x_{\wh j_{t,k-1}}) = {\bar c}(x \mid x_{\wh j_{t,1}}, \ldots, x_{\wh j_{t,k-1}})^\top M^{-1}_{c_t-1} {\bar c}(x \mid x_{\wh j_{t,1}}, \ldots, x_{\wh j_{t,k-1}}) \,\alpha(b,d',T,\delta)$ 
    \item Let ${\wh J_{t,s}} = \langle x_{\wh j_{t,1}}, \ldots, x_{\wh j_{t,s}} \rangle$ for any $s \le b_t$;
    \vspace{-0.05in}
\item Set\ \ 
\(
\displaystyle{{\wh s_t} = \arg\max_{s=0,1,\ldots,b_t} {\wh \E_{Y_t}}[R({\wh J_{t,s}},Y_t)]}~,
\)
with 
\vspace{-0.05in}
\begin{eqnarray*}
\hDelta_{{\wh j_{t,k}},t} 
&=& 
{\bar c}(x_{\wh j_{t,k}} \mid x_{\wh j_{t,1}}, \ldots, x_{\wh j_{t,k-1}})^\top w_{c_t}\\
\epsilon_{{\wh j_{t,k}},t}^2 
&=& 
{\bar c}(x_{\wh j_{t,k}} \mid x_{\wh j_{t,1}}, \ldots, x_{\wh j_{t,k-1}})^\top M^{-1}_{c_t-1}\,{\bar c}(x_{\wh j_{t,k}} \mid x_{\wh j_{t,1}}, \ldots, x_{\wh j_{t,k-1}})\ \alpha(b,d',T,\delta)\\
{\wh \E_{Y_t}}[R({\wh J_{t,s}},Y_t)] 
&=& 
\begin{cases}
E\left(\hDelta_{{\wh j_{t,1}},t}+\epsilon_{{\wh j_{t,1}},t},\ldots,\hDelta_{{\wh j_{t,s}},t}+\epsilon_{{\wh j_{t,s}},t}\right) &{\mbox{if $s \geq 1$}}\\
\ell_{0,t} &{\mbox{otherwise}}~,
\end{cases}
\end{eqnarray*}
where function $E(\cdot,\ldots,\cdot)$ is as in (\ref{e:ehat}) in Lemma \ref{l:monotonicity}, with $p(\cdot)$ therein replaced by $\sigma(\cdot)$~;
\item Finally, $J_t = {\wh J_{t,{\wh s_t}}}$~;
\end{itemize}
\vspace{-0.05in}
\item Observe feedback $Y_t \downarrow J_t = 
\begin{cases}
\langle y_{t,{\wh j_{t,1}}}, y_{t,{\wh j_{t,2}}}, \ldots, y_{t,{\wh j_{t,{\wh s'_t}}}}\rangle = \langle 0,\ldots, 0, 1\rangle, &{\mbox{for some ${\wh s'_t} \leq {\wh s_t}$\qquad or}}\\
\langle y_{t,{\wh j_{t,1}}}, y_{t,{\wh j_{t,2}}}, \ldots, y_{t,{\wh j_{t,{\wh s_t}}}}\rangle = \langle 0,\ldots, 0, 0\rangle
\end{cases}
$
\item {\bf For} $k = 1,\ldots, {\wh s_t}$ (in the order of occurrence of items in $J_t$)
update :
\begin{align*}
    M_{c_t+k-1} &= M_{c_t+k-2} + |s_{k,t}|\ {\bar c}(x_{\wh j_{t,k}} \mid x_{\wh j_{t,1}}, \ldots, x_{\wh j_{t,k-1}}) \ {\bar c}(x_{\wh j_{t,k}} \mid x_{\wh j_{t,1}}, \ldots, x_{\wh j_{t,k-1}})^\top\,,\\
    w_{c_t+k} &= w'_{c_t+k-1} + \frac{1}{c_{\sigma'}} M^{-1}_{c_t+k-1}\nabla_{k,t}~,
\end{align*}
where
\vspace{-0.05in}
\[
s_{k,t} =
\begin{cases}
1 & {\mbox{If $y_{t,k}$ is observed and $y_{t,k}= 1$}}\\
-1 & {\mbox{If $y_{t,k}$ is observed and $y_{t,k}= 0$}}\\
0 & {\mbox{If $y_{t,k}$ is not observed}}~,
\end{cases}
\]
and
\(
\nabla_{k,t} = \sigma(-s_{k,t}\,\hDelta'_{k,t})\,s_{k,t}\,{\bar c}(x_{\wh j_{t,k}} \mid x_{\wh j_{t,1}}, \ldots, x_{\wh j_{t,k-1}})~,
\)
where $\hDelta'_{k,t} = {\bar c}(x_{\wh j_{t,k}} \mid x_{\wh j_{t,1}}, \ldots, x_{\wh j_{t,k-1}})^\top w'_{c_t+k-1}$\\ 
with
$$
w'_{c_t+k-1} = \arg\min_{w\,:\,-D \leq w^\top {\bar c}(x_{\wh j_{t,k}} \mid x_{\wh j_{t,1}}, \ldots, x_{\wh j_{t,k-1}}) \leq D} \ d_{c_t+j-2}(w,w_{c_t+k-1})~;
$$
\vspace{-0.05in}
\item $c_{t+1} \leftarrow c_t + {\wh s_t}$~.
\end{enumerate}
\caption{\label{a:dependent}The contextual bandit algorithm in the dependent case. Here the link function $\sigma(\cdot)$ is $\sigma(x) = \frac{\exp(x)}{1+\exp(x)}$.}
\end{algorithm*}

Below we give the derivation for the approximation ratio claimed in the main body of the paper
\begin{lemma}\label{l:approx}
Fix $s \in \{0,1,\ldots, b\}$. Let $J^* = \langle x_{j_1^*},\ldots, x_{j_s^*} \rangle$ be the Bayes optimal sequence under model (\ref{e:dependent_model}) with unknown vector $u$. Let $(x_{j_1'},x_{j_2'},\ldots)$ be the order of items according to Eq.\eqref{eq:greedy} and the unknown vector $u$ and $J' = \langle x_{j_1'},\ldots, x_{j_s'} \rangle$ be the sequence taking first $s$ elements.
Suppose ${\bar c}(x_k\,|\,x_1,\ldots,x_{k-1} )^\top u \in [-D,D]$ for all $x_1, \ldots, x_k \in A$. Assume all components of $u$ are non-negative and that\footnote
{
Notice that, since $z= 1$ and $c_{\sigma'} = e^{-D}/(1+e^{-D})^2$, this requirement is essentially equivalent to something like $||u||_1 = O(\sqrt{d'})$.
}
$\|u\|_1 \le \frac{\sqrt{d'}(z-(1-1/e)c_{\sigma'})}{6(z^2 - (1-1/e)c_{\sigma'}^2)} $.
%
%
Moreover, let the reward and loss sequences satisfy\footnote
{
For example, this requirement holds when $r_s \ge 5 |\ell_s|$ for all $s \geq 1$, and $\frac{c_{\sigma'}}{z} \le \frac{1}{2(1-1/e)}$.
}
\begin{align*}
    s (r_s - \ell_s) \max&\left\{\frac{1}{s}, 1 - \frac{s-1}{2}c_{\sigma}\right\} \left(1 - \left(1 - \frac{1}{e}\right)\frac{c_{\sigma'}}{z}\right)\\ &+ 3\ \ell_s \left(1 - \max\left\{\frac{1}{s}, 1 - \frac{s-1}{2}c_{\sigma}\right\} \left(1 - \frac{1}{e}\right) \frac{c_{\sigma'}(r_s - \ell_s)}{z(r_1 - \ell_s)}\right) \ge 0\,.
\end{align*}
Let
\begin{align*}
    \gamma(s) = 
    \begin{cases}
        \max\{\frac{1}{s}, 1 - \frac{s-1}{2}c_{\sigma}\} (1 - \frac{1}{e}) \frac{c_{\sigma'}(r_s - \ell_s)}{z(r_1 - \ell_s)} &s\ge 2\,,\\
        1 &s=0,1~.
\end{cases}
\end{align*}
Then 
\begin{align*}
    \E_{Y}[R(J',Y)] \ge \gamma(s)\ \E_{Y}[R(J^\ast,Y)]\,.
\end{align*}
\end{lemma}
\begin{proof}\allowdisplaybreaks
It is immediate to see the conclusion holds for $s=0,1$. Now assume $s \ge 2$. Let $J = \langle x_{j_1},\ldots, x_{j_s} \rangle$ be any sequence of length $s$. Then, setting for brevity $a= 2/\sqrt{d'}$ and $a' = -1/\sqrt{d'}(1,\ldots,1)^\top$,
we can write
\begin{align*}
        \E_{Y}[R(J,Y)] &= E(\Delta_{j_1}, \Delta_{j_1,j_2}, \ldots, \Delta_{j_1,j_2,\ldots,j_s})\\
        &= r_1 p(\Delta_{j_1}) + r_2 p(\Delta_{j_1,j_2}) (1 - p(\Delta_{j_1})) + \cdots + r_s p(\Delta_{j_1,\ldots,j_s}) \prod_{i=1}^{s-1}(1 - p(\Delta_{j_1,\ldots,j_i}))\\ 
        &\hspace{3in}+ \ell_s (1 - \prod_{i=1}^{s}(1 - p(\Delta_{j_1,\ldots,j_i})))\\
        &= (r_1-\ell_s) p(\Delta_{j_1}) + (r_2-\ell_s) p(\Delta_{j_1,j_2}) (1 - p(\Delta_{j_1})) + \cdots\\ 
        &\hspace{1.95in}+ (r_s-\ell_s) p(\Delta_{j_1,\ldots,j_s}) \prod_{i=1}^{s-1}(1 - p(\Delta_{j_1,\ldots,j_i})) + \ell_s\\
        &\le (r_1-\ell_s) (1 - \prod_{i=1}^{s}(1 - p(\Delta_{j_1,\ldots,j_i}))) + \ell_s\\
        &\le (r_1-\ell_s) \sum_{i=1}^{s} p(\Delta_{j_1,\ldots, j_i}) + \ell_s\\
        &= (r_1-\ell_s) \sum_{i=1}^{s} \sigma(a\cdot c'(x_{j_i} \mid x_{j_1}, \ldots, x_{j_{i-1}})^\top u + a'^\top u) + \ell_s \\
        &\le (r_1-\ell_s) \sum_{i=1}^{s} \left(\sigma(a\cdot c'(x_{j_i} \mid x_{j_1}, \ldots, x_{j_{i-1}})^\top u) + c_{\sigma'}a'^\top u \right) + \ell_s\\
        &\le (r_1-\ell_s) \sum_{i=1}^{s} \left(\sigma(0) + z\cdot a\cdot c'(x_{j_i} \mid x_{j_1}, \ldots, x_{j_{i-1}})^\top u + c_{\sigma'}a'^\top u \right) + \ell_s \\
        &= (r_1-\ell_s) (s/2 + s\ c_{\sigma'}\ a'^\top u + z\cdot a\cdot \langle c'(\{x_{j_1}, \ldots, x_{j_{s}}\}), u \rangle) + \ell_s\\
        &= (r_1-\ell_s) (s/2 + s\ c_{\sigma'}\ a'^\top u + z\cdot a\cdot \langle c'(J), u \rangle) + \ell_s\,
    \end{align*}
    where the fourth and third lines from last are both from the properties of the $\sigma$ function.
    Also
    \begin{align*}
        \E_{Y}&[R(J,Y)]\\ 
        &\ge (r_s - \ell_s) (1 - \prod_{i=1}^{s}(1 - p(\Delta_{j_1,\ldots,j_i}))) + \ell_s\\
        &\ge (r_s - \ell_s) \max\Bigl\{\frac{1}{s}, 1 - \frac{s-1}{2}c_{\sigma}\Bigl\} \sum_{i=1}^{s} p(\Delta_{j_1,\ldots, j_i}) + \ell_s\\
        &= (r_s - \ell_s) \max\Bigl\{\frac{1}{s}, 1 - \frac{s-1}{2}c_{\sigma}\Bigl\} \sum_{i=1}^{s} \sigma(a\cdot c'(x_{j_i} \mid x_{j_1}, \ldots, x_{j_{i-1}})^\top u + a'^\top u) + \ell_s \\
        &\ge (r_s - \ell_s) \max\Bigl\{\frac{1}{s}, 1 - \frac{s-1}{2}c_{\sigma}\Bigl\} \sum_{i=1}^{s} \left( \sigma(a\cdot c'(x_{j_i} \mid x_{j_1}, \ldots, x_{j_{i-1}})^\top u) + z\cdot a'^\top u \right) + \ell_s\\
        &\ge (r_s - \ell_s) \max\Bigl\{\frac{1}{s}, 1 - \frac{s-1}{2}c_{\sigma}\Bigl\} \sum_{i=1}^{s} \left( \sigma(0) + c_{\sigma'}\ a\cdot c'(x_{j_i} \mid x_{j_1}, \ldots, x_{j_{i-1}})^\top u + z\cdot a'^\top u \right) + \ell_s\\
        &= (r_s - \ell_s) \max\Bigl\{\frac{1}{s}, 1 - \frac{s-1}{2}c_{\sigma}\Bigl\} (s/2 +s\ z\ a'^\top u+ c_{\sigma'}\ a\ \langle c'(\{x_{j_1}, \ldots, x_{j_{s}}\}), u\rangle ) + \ell_s\\
        &=(r_s - \ell_s) \max\Bigl\{\frac{1}{s}, 1 - \frac{s-1}{2}c_{\sigma}\Bigl\} (s/2 +s\ z\ a'^\top u+ c_{\sigma'}\ a\ \langle c'(J), u\rangle ) + \ell_s\,,
    \end{align*}
    where the second inequality is by Lemma 1 of \cite{h+20}, and the fourth and fifth lines are from the properties of the $\sigma$ function. Thus
\begin{align*}
\E_{Y}&[R(J',Y)] 
\ge 
(r_s - \ell_s) \max\Bigl\{\frac{1}{s}, 1 - \frac{s-1}{2}c_{\sigma}\Bigl\} \left(s/2 + s\ z\ a'^\top u + a\  c_{\sigma'}\langle c'(J'), u\rangle\right) + \ell_s\\
&\ge 
(r_s - \ell_s) \max\Bigl\{\frac{1}{s}, 1 - \frac{s-1}{2}c_{\sigma}\Bigl\} \left(s/2 + s\ z\ a'^\top u + a\ c_{\sigma'}\Bigl(1 - \frac{1}{e}\Bigl)\,\max_{J} \langle c'(J), u\rangle\right) + \ell_s\\
&\ge 
(r_s - \ell_s) \max\Bigl\{\frac{1}{s}, 1 - \frac{s-1}{2}c_{\sigma}\Bigl\} \left(s/2 + s\ z\ a'^\top u + a\ c_{\sigma'}\Bigl(1 - \frac{1}{e}\Bigl)\langle c'(J^*), u\rangle\right) + \ell_s\\
&\ge 
(r_s - \ell_s) \max\Bigl\{\frac{1}{s}, 1 - \frac{s-1}{2}c_{\sigma}\Bigl\} \\
&\hspace{0.1in}\times\left(s/2 + s\ z\ a'^\top u + a\ c_{\sigma'}\Bigl(1 - \frac{1}{e}\Bigl) \left( \frac{\E_{Y}[R(J^*,Y)] - \ell_s}{a\ z\ (r_1 - \ell_s)} - \frac{s}{2az} - \frac{s\ c_{\sigma'}\ a'^\top u}{az} \right) \right) + \ell_s\\
&\ge 
\max\Bigl\{\frac{1}{s}, 1 - \frac{s-1}{2}c_{\sigma}\Bigl\} \Bigl(1 - \frac{1}{e}\Bigl) \frac{c_{\sigma'}(r_s - \ell_s)}{z(r_1 - \ell_s)} \E_{Y}[R(J^*,Y)]\\
&
\quad + (r_s - \ell_s) \max\Bigl\{\frac{1}{s}, 1 - \frac{s-1}{2}c_{\sigma}\Bigl\} s \left(\frac{1}{2} (1 - \Bigl(1 - \frac{1}{e}\Bigl)\frac{c_{\sigma'}}{z}) + a'^\top u (z - \Bigl(1 - \frac{1}{e}\Bigl)\frac{c_{\sigma'}^2}{z}) \right) \\
&
\quad + \ell_s \left(1 - \max\Bigl\{\frac{1}{s}, 1 - \frac{s-1}{2}c_{\sigma}\Bigl\} \Bigl(1 - \frac{1}{e}\Bigl) \frac{c_{\sigma'}(r_s - \ell_s)}{z(r_1 - \ell_s)}\right)\\
&\ge 
\max\Bigl\{\frac{1}{s}, 1 - \frac{s-1}{2}c_{\sigma}\Bigl\} \Bigl(1 - \frac{1}{e}\Bigl) \frac{c_{\sigma'}(r_s - \ell_s)}{z(r_1 - \ell_s)} \E_{Y}[R(J^*,Y)]~.
\end{align*}
In the above, the second inequality is based on the fact that the selection of $J'$ is equivalent to running {\sc greedy} on maximizing $\langle c(J),u \rangle$ over $J$, along with the typical approximation ratio of monotone and sub-modular set function optimization. The third inequality is by $\max_{J} \langle c'(J), u\rangle \ge \langle c'(J^*), u\rangle$. The fourth inequality is by the lower bound of $\E_{Y}[R(J^*,Y)]$ in terms of $\langle c'(J^*), u\rangle$. The last inequality is by the definition of $a'$, and the assumptions on $r_s,\ell_s$.
\end{proof}

The next lemma is the dependent outcome counterpart to Lemma \ref{l:onetimeregret}.
%
%
\begin{lemma}\label{l:onetimeregret dependent}
Let us assume the dependent model (\ref{e:dependent_model}) for outcome vector $Y$. Then, for given set of actions $A$, and budget $b$, let $J^*$ be the Bayes optimal sequence and $J = \langle x_{j_1},\ldots,x_{j_{s}} \rangle$ be the sequence computed by Algorithm \ref{a:dependent} on $A$ and $b$, with link function $\sigma$ such that $\sigma'(\Delta) \leq z$ for all $\Delta \in \R$. Further, let $\Delta_{j_1, \ldots, j_k} = u^\top {\bar c}(x_{j_k} \mid x_{j_1}, \ldots, x_{j_{k-1}})$, and $\hDelta_{j_1, \ldots, j_k} = w^\top {\bar c}(x_{j_k} \mid x_{j_1}, \ldots, x_{j_{k-1}})$, for all conditional vectors computed from $A$, and assume $|\Delta_{j_1, \ldots, j_k} - \hDelta_{j_1, \ldots, j_k}| \leq \epsilon_{j_1, \ldots, j_k}$ for all $j$ sequence, where $w$ is the vector used by Algorithm \ref{a:dependent} to compute $J$. 
Suppose\footnote{
This requirement is controllable since $\epsilon_{j_1,\ldots,j_k}$ is reasonably small after $O(\log T)$ rounds.
}
$\Delta_{j_1,\ldots,k_k} + 2\epsilon_{j_1,\ldots,j_k} \in [-D,D]$ for all $x_{j_1}, \ldots, x_{j_k} \in A$.
Then the scaled one-time regret (\ref{e:onetimescaled}) can be bounded as follows:
\begin{align*}
\E_{Y} [\gamma(s_t^\ast) R(J^*,Y)] - \E_{Y} [R(J,Y)]\leq 
\begin{cases}
4z\,\sum_{i = 1}^s \epsilon_{j_1,\ldots,j_i}\,\prod_{h=1}^{i-1}
(1-\sigma(\Delta_{j_1,\ldots,j_h})) &{\mbox{if $J \neq \langle\rangle$}}\\
0 &{\mbox{otherwise}}~.
\end{cases}
\end{align*}
\end{lemma}
\begin{proof}
Irrespective of whether $J \neq \langle\rangle$ or  $J^* \neq \langle\rangle$, we can write
\begin{align*}
\E_{Y}&[\gamma(s_t^\ast) R(J^*,Y)] - \E_{Y} [R(J,Y)] \\
    &\le {\wh \E_{Y}}[\gamma(s_t^\ast) R(J^*,Y)] - \E_{Y} [R(J,Y)]\\
    &\le {\wh \E_{Y}}[R(J',Y)] - \E_{Y} [R(J,Y)]\\
    &\le {\wh \E_{Y}}[R(J,Y)] - \E_{Y} [R(J,Y)]\\
    &=E(\hDelta_{j_1}+\epsilon_{j_1},\hDelta_{j_1,j_2}+\epsilon_{j_1,j_2}, \ldots, \hDelta_{j_1,j_2,\ldots,j_s}+\epsilon_{j_1,j_2,\ldots,j_s})
    - E(\Delta_{j_1},\Delta_{j_1,j_2}\ldots, \Delta_{j_1,j_2,\ldots,j_s}) \,,
\end{align*}
where ${\wh \E_{Y}}$ is defined as in Algorithm \ref{f:2} by using $\hDelta_{j_1,\ldots,k_k} + \epsilon_{j_1,\ldots,j_k}$. Here $J'$ is computed similarly in Lemma \ref{l:approx} but under $\hDelta_{j_1,\ldots,k_k} + \epsilon_{j_1,\ldots,j_k}$ and length $s_t^\ast$. The $\gamma(s_t^\ast)$-approximation still holds according to Lemma \ref{l:approx}.
The list $J'$ is just $\hat{J}_{s_t^\ast}$ in Algorithm \ref{a:dependent} and is no better than $J$ under ${\wh \E_{Y}}$ according to the computation of $s$.

Similar to the proof of Lemma \ref{l:onetimeregret}, by the mean-value theorem, we can write
\begin{align*}
    E(\Delta_{j_1}+2\epsilon_{j_1},&\Delta_{j_1,j_2}+2\epsilon_{j_1,j_2}, \ldots, \Delta_{j_1,j_2,\ldots,j_s}+2\epsilon_{j_1,j_2,\ldots,j_s}) - E(\Delta_{j_1},\Delta_{j_1,j_2}\ldots, \Delta_{j_1,j_2,\ldots,j_s})\\
    =&2\,\sum_{i = 1}^s \frac{\partial E(\Delta_{j_1},\Delta_{j_1,j_2}\ldots, \Delta_{j_1,j_2,\ldots,j_s})}{\partial \Delta_{j_1,j_2,\ldots,j_i}}{\big|}_{\Delta_{j_1} = \xi_{j_1},\ldots,\Delta_{j_1,\ldots,j_s} = \xi_{j_1,\ldots,j_s} }\, \epsilon_{j_1,\ldots,j_i}\,,
\end{align*}
where $\xi_{j_1,\ldots,j_i} \in (\Delta_{j_1,\ldots,j_i},\Delta_{j_1,\ldots,j_i}+2\epsilon_{j_1,\ldots,j_i})$.
The third part of Lemma \ref{l:monotonicity} then allows us to write
\begin{align*}
\frac{\partial E(\Delta_{j_1},\Delta_{j_1,j_2}\ldots, \Delta_{j_1,j_2,\ldots,j_s})}{\partial \Delta_{j_1,j_2,\ldots,j_i}}&{\big|}_{\Delta_{j_1} = \xi_{j_1},\ldots,\Delta_{j_1,\ldots,j_s} = \xi_{j_1,\ldots,j_s} }\\ 
&\hspace{1in}\leq 2z\, (1-\sigma(\xi_{j_1}))\cdots (1-\sigma(\xi_{j_1,\ldots,j_{i-1}})) \\
&\hspace{1in}\leq 2z\, (1-\sigma(\Delta_{j_1}))\cdots (1-\sigma(\Delta_{j_1,\ldots,j_{i-1}}))~,
\end{align*}
the second inequality deriving from the monotonicity of $\sigma(\cdot)$ and the fact that $\xi_{j_1,\ldots,j_i} \in (\Delta_{j_1,\ldots,j_i},\Delta_{j_1,\ldots,j_i}+2\epsilon_{j_1,\ldots,j_i})$.
Replacing back, and summing over $i$ yields the claimed bound.
\end{proof}

Based on this lemma, we combine with the corresponding remaining parts in the proof for the independent case. This gives us a {\em scaled} regret bound which coincides with the one for the dependent case.

Yet, it is worth stressing that, despite the two regret {\em bounds} look alike, the two underlying notions of regret are widely different, both because we have now a scaled regret, and because of the different assumptions on the process generating the outcomes as compared to the independent case.

\section{Further Related Work}\label{as:further}

\cite{kveton2015combinatorial} studies a variant of cascading bandits where the feedback stops when a $0$ outcome is observed, as opposed to a $1$ outcome of the standard cascading bandit model. This reward is equivalent to a Boolean AND function on the sequence, and the available sequences are defined by combinatorial constraints of the problem. \cite{zhou2018cost} also studies a variant of cascading bandits where each arm has an extra (unknown) cost when displayed. The length of the recommended sequences can also change, but in their setting this is due to the trade-off between the attractiveness and the cost of an item, while in our setting this is due to the trade-off between attractiveness of items and both reward and loss values. 
The
combinatorial
semi-bandit setting with probabilistically triggered arms \cite{wc17} is a generalization of the cascading bandit setting that also encompasses, for instance, influence maximization problems. The authors are able to remove the inconvenient dependence on $1/p^*$ alluded to at the end of Section \ref{s:independent}, but their comprehensive analysis only applies to non-contextual bandit scenarios.

Besides cascading bandits, relevant works investigate bandits with submodular reward functions to account for diversity in the item assortment (e.g., \cite{yg11,takemori2020submodular}). In particular, \cite{takemori2020submodular} 
show a regret bound of the form $\sqrt{bT}$ in a submodular bandits scenario with rewards on items similar to our setting, yet relying on a feedback which is more informative than ours. For instance, in the independent case, their setting is equivalent to a (constrained) combinatorial bandits scenario with {\em semi-bandit} feedback with linear rewards.

Regarding the generative model for outcome vectors, following previous work \citep{lz18}, we assumed the probability that an item is successful is ruled by a generalized linear model (GLM), Such a model is more convenient than a purely linear model, since the sigmoidal link function would always map values to $(0,1)$ which we need here to encode probabilities and compute the Bayes optimal sequence. 
The bandit problem under GLM assumptions is first studied in \cite{f+10}, whose regret bound can be improved by the finer self-concordant analysis of \cite{DBLP:journals/corr/abs-2002-07530}. 
The online Newton step analysis presented here is inspired by the GLM-based bandit analysis contained in \cite{DBLP:journals/corr/abs-1207-0166}. See also \cite{zhang2016online} for similar results.
\cite{li2017provably} gives an optimal solution for this model up to a constant coefficient.

Finally, the update method that deals with long sequences in our paper also often appears in the study of bandit algorithms with delayed feedback. There is indeed some kind of similarity between a cascading model and a delayed feedback model in bandits: both share the need for a bandit algorithm to deal with signals that are received somehow later than the time the algorithm commit to actions.
Relevant works in bandits with delayed feedback include
\cite{dudik2011efficient, joulani2013online,JMLR:v20:17-631,pike2018bandits,zhou2019learning,arya2020randomized}. Yet, we are not aware of a way to reduce the delayed bandit model to the cascading bandit model, or vice versa.

\section{Further Experimental Results}\label{s:furtherexp}

This section contains details on our experimental setting and results that have been omitted from the main paper.

\subsection{Dataset Preprocessing} 

We report here the pre-processing steps we followed for the Million Songs, Yelp, and MNIST datasets.

\begin{itemize}
\item {\bf Million Songs:} The Million Songs Dataset (MSD) is a repository of audio features and metadata of a million contemporary pop songs. We consider the Echo Nest Taste Profile Subset of MSD that contains the play-counts of some of these songs by real users. We pick $100,000$ users that have played the highest number of songs and $50,000$ songs with the highest number of users. We sample $10,000$ songs at random and calculate the singular value decomposition (SVD) of the corresponding $100,000 \times 10,000$ ratings matrix into $10$ principal components. The projection matrices from the SVD are used to compute embeddings of dimension $d=10$ for the remaining $40,000$ songs for training the bandit algorithms. The embeddings are normalized to unit $L_2$-norm and the dataset is shuffled randomly. In every round of bandit learning, the algorithm is presented with a non-overlapping chunk of movies as arms ($A_t$). The chunk size is $100$.
The rate of success of an arm is decided by the mean rating received by the corresponding movie in the dataset. This mean rating is normalized by first re-centering through its median value in the dataset, and then converting to a probability by passing through a sigmoidal function.
As mentioned in Section \ref{ss:param}, for the dependent algorithm the 40,000 SVD-projected $d$-dimensional vectors have been used to compute coverage vectors through a Gaussian Mixture Model (GMM) with $d'$ centroids.

\item {\bf Yelp:} The Yelp Dataset Challenge is a library of restaurants (and related businesses) and their reviews from customers. We pick $200,000$ users that have reviewed the highest number of businesses and $50,000$ businesses with the highest number of reviews. We sample $10,000$ businesses at random and calculate the singular value decomposition (SVD) of the corresponding $200,000 \times 10,000$ ratings matrix into $10$ principal components. The projection matrices from the SVD are used to compute embeddings of dimension $d=10$ for the remaining $40,000$ businesses for training the bandit algorithms. The embeddings are normalized to unit $L_2$-norm and the dataset is shuffled randomly. In every round of bandit learning, the algorithm is presented with a non-overlapping chunk of movies as arms ($A_t$). The chunk size is $100$.
The rate of success of an arm is decided by the mean rating received by the corresponding movie in the dataset. This mean rating is normalized by first re-centering through its median value in the dataset, and then converting to a probability by passing through a sigmoidal function.
As mentioned in Section \ref{ss:param}, for the dependent algorithm the $40,000$ SVD-projected $d$-dimensional vectors have been used to compute coverage vectors through a GMM with $d'$ centroids.
\item {\bf MNIST:} The MNIST dataset consists of $60,000$ training samples and $10,000$ test samples. We draw $19,800$ samples at random from the training split for constructing a $d=10$-dimensional embedding space using Principal Component Analysis (PCA) and combine the remaining training samples with the test samples and randomly shuffle it to create a dataset of $50,200$ samples for training the bandit algorithm. As mentioned in Section \ref{ss:param}, for the dependent algorithm the $50,200$ SVD-projected $10$-dimensional vectors are used to compute coverage vectors through a GMM with $d'$ centroids.  All observed vectors (embeddings and coverage vectors) are scaled to unit $L_2$-norm.

MNIST has $10$ output classes. For each of these output classes, we define a sub-task that considers that class as the ``pivot-class''. At every round of bandit learning, we present the agent with a non-overlapping chunk of examples as arms. The agent observes success only if it chooses an arm whose output class matches the pivot class. We choose the pivot class at the beginning of each experiment and keep it constant throughout. 
\end{itemize}

\subsection{Metric} 

We report the algorithms' performance in terms of their Normalized Cumulative Reward ($NCR$) obtained over all rounds of training. If a given dataset has $T$ chunks then each algorithm is trained for exactly $T$ rounds. The Cumulative Reward ($CR$) obtained by an algorithm is normalized with respect to the $CR$ accrued by the random policy and the maximum possible $CR$ over $T$ rounds with budget $b_t$ in each round to obtain $NCR$ as:
\begin{equation*}
    NCR = \frac{CR\_algorithm - CR\_random}{CR\_max - CR\_random}~.
\end{equation*}
$NCR$ is meant to capture the fractional advantage in $CR$ over the random policy Rand. This normalization is needed especially when the random policy shows good performance, for example, in the vanilla scenario with large $b_t$.

\subsection{Results} 
Our $NCR$ results are reported in Tables
\ref{tab:msd-ncr-full-msd}--\ref{tab:msd-ncr-full-mnist9}. Notice that, by construction, the $NCR$ of Rand is always 0.0. Also observe that for large $b_t$, the vanilla scenario makes all algorithms essentially indistinguishable, and when $b_t = 50$ or $b_t = 100$ also Rand performs as well as all other algorithms. This is not the case for the exponential scenario. In a few cases, the tables reflect negative entries (specifically for Eps), which simply means that the algorithm happened to perform worse than Rand.

From these experiments, the following trends emerge. 
\begin{enumerate}
    \item In a vanilla scenario that emphasizes early success ($b_t$ small), the baseline algorithms (Eps, C-UCB1, GL-CDCM) are rarely the winner. In most cases, the winner is either the proposed independent (Ind) or dependent (Dep) algorithms. On the other hand, as the budget $b_t$ grows the algorithms tend to be indistinguishable. This has to be expected, as when $b_t$ is large even the random policy (Rand) becomes competitive in the vanilla scenario, and $NCR$s, by their very definition, tend to be zero.
    \item In the exponential scenario, Dep generally outperforms Ind, with the exception of a few cases in the MNIST dataset (specifically on pivot classes $3$, $4$, $5$, $8$, and $9$). For these tasks, Dep is dramatically underperforming, probably due to the latent space construction, which does not offer a convenient representation -- see Section \ref{ssa:further}.
\end{enumerate}

\begin{table}[t!]
\smaller
\centering
\caption{Comparison of normalized cumulative reward on the Million Songs Dataset for two different reward/loss scenarios -- Vanilla and Exponential. ``Rand" refers to the random policy, ``Eps" is the $\epsilon$-greedy version of our Algorithm \ref{f:2}, ``C-UCB1" is the cascading bandit algorithm of \cite{z+16}, while ``GL-CDCM" is the one from \cite{liu2018contextual}. Moreover, ``Ind" and ``Dep" are abbreviations for the Independent (Algorithm \ref{f:2}) and Dependent (Algorithm \ref{a:dependent}) algorithms proposed in this paper. Notice that the exponential scenario does not include the baselines ``Eps", ``C-UCB1" and ``GL-CDCM" since those baselines are defined to work only in the vanilla scenario. For each of the two scenarios and each value of $b_t$, we emphasize in bold the best performance.}
\label{tab:msd-ncr-full-msd}
\begin{tabular}{c|c|c|c|c|c|c||c|c|c|}
\cline{2-10}
\multicolumn{1}{l|}{}               & \multicolumn{6}{c||}{Vanilla} & \multicolumn{3}{c|}{Exponential} \\ \hline
\multicolumn{1}{|c|}{$b_t$}  & Rand     & Eps         & C-UCB1  & GL-CDCM   & Ind         & Dep     &Rand     & Ind       & Dep                \\ \hline
\multicolumn{1}{|c|}{1}      & 0.00     & 0.27 &       0.29 &  0.28 &      0.32 &       {\bf 0.51}      & 0.00    &  0.32 &       {\bf 0.48}                      \\ \hline
\multicolumn{1}{|c|}{5}      & 0.00     & 0.40 &       0.50 &  {\bf 0.90} &      {\bf 0.90} &       {\bf 0.90}      & 0.00    &  0.35 &       {\bf 0.56}                        \\ \hline
\multicolumn{1}{|c|}{10}     & 0.00     & {\bf 0.15} &       {\bf 0.15} &  {\bf 0.15} &      {\bf 0.15} &       {\bf 0.15}      & 0.00    &  0.38 &       {\bf 0.57}                       \\ \hline
\multicolumn{1}{|c|}{50}     & {\bf 0.00}     & {\bf 0.00} &       {\bf 0.00} &  {\bf 0.00} &      {\bf 0.00} &       {\bf 0.00}      & 0.00    &  0.38 &       {\bf 0.50}                        \\ \hline
\multicolumn{1}{|c|}{100}    & {\bf 0.00}     & {\bf 0.00} &      {\bf 0.00} &  {\bf 0.00} &     {\bf 0.00} &       {\bf 0.00}    & 0.00    &  0.34 &      {\bf 0.54}                       \\ \hline
\end{tabular}
\end{table}

\begin{table}[t!]
\smaller
\centering
\caption{Same as in Table \ref{tab:msd-ncr-full-msd} for the Yelp dataset.}
\label{tab:msd-ncr-full-yelp}
\begin{tabular}{c|c|c|c|c|c|c||c|c|c|}
\cline{2-10}
\multicolumn{1}{l|}{}               & \multicolumn{6}{c||}{Vanilla} & \multicolumn{3}{c|}{Exponential} \\ \hline
\multicolumn{1}{|c|}{$b_t$}  & Rand     & Eps         & C-UCB1  & GL-CDCM   & Ind         & Dep     &Rand     & Ind       & Dep                \\ \hline
\multicolumn{1}{|c|}{1}      & 0.00     & 0.17 &       0.06 &  0.16 &      0.16 &       {\bf 0.25}      & 0.00    &  0.18 &       {\bf 0.26}                      \\ \hline
\multicolumn{1}{|c|}{5}      & 0.00     & 0.33 &       0.53 &  0.67 &      0.67 &       {\bf 0.80}      & 0.00    &  0.17 &       {\bf 0.31}                        \\ \hline
\multicolumn{1}{|c|}{10}     & 0.00     & {\bf 0.35} &       {\bf 0.35} &  {\bf 0.35} &      {\bf 0.35} &       {\bf 0.35}      & 0.00    &  0.23 &       {\bf 0.28}                        \\ \hline
\multicolumn{1}{|c|}{50}     & {\bf 0.00}     & {\bf 0.00} &       {\bf 0.00} &  {\bf 0.00} &      {\bf 0.00} &       {\bf 0.00}      & 0.00    &  0.21 &       {\bf 0.26}                        \\ \hline
\multicolumn{1}{|c|}{100}    & {\bf 0.00}     & {\bf 0.00} &       {\bf 0.00} &  {\bf 0.00} &      {\bf 0.00} &       {\bf 0.00}    & 0.00    &  0.18 &       {\bf 0.29}                        \\ \hline
\end{tabular}
\end{table}

\begin{table}[t!]
\smaller
\centering
\caption{Same as in Table \ref{tab:msd-ncr-full-msd} for the Movielens dataset.}
\label{tab:movielens-ncr-full-movielens}
\begin{tabular}{c|c|c|c|c|c|c||c|c|c|}
\cline{2-10}
\multicolumn{1}{l|}{}               & \multicolumn{6}{c||}{Vanilla} & \multicolumn{3}{c|}{Exponential} \\ \hline
\multicolumn{1}{|c|}{$b_t$}  & Rand     & Eps         & C-UCB1  & GL-CDCM   & Ind         & Dep     &Rand     & Ind       & Dep                \\ \hline
\multicolumn{1}{|c|}{1}      & 0.00     & 0.26 &       0.17    &  0.35 &     {\bf 0.40} &       0.34      & 0.00    &  0.35 &       {\bf 0.39}                      \\ \hline
\multicolumn{1}{|c|}{5}      & 0.00     & 0.23 &       0.51 &  0.78 &      0.62 &       {\bf 0.84}      & 0.00    & 0.36 &       {\bf 0.43}                        \\ \hline
\multicolumn{1}{|c|}{10}     & 0.00     & {\bf 0.38} &       {\bf 0.38} &  {\bf 0.38} &      {\bf 0.38} &       {\bf 0.38}      & 0.00    & 0.35 &       {\bf 0.38}                        \\ \hline
\multicolumn{1}{|c|}{50}     & {\bf 0.00}    & {\bf 0.00} &       {\bf 0.00} &  {\bf 0.00} &      {\bf 0.00} &       {\bf 0.00}      & 0.00    & {\bf 0.40} &       0.39                        \\ \hline
\multicolumn{1}{|c|}{100}    & {\bf 0.00}     &{\bf 0.00} &       {\bf 0.00} &  {\bf 0.00} &     {\bf 0.00} &       {\bf 0.00}    & 0.00    & 0.32 &       {\bf 0.38}                       \\ \hline
\end{tabular}
\end{table}

\begin{table}[t!]
\smaller
\centering
\caption{
Comparison of normalized cumulative reward on the MNIST Dataset with pivot-class $0$ for two different reward/loss scenarios -- Vanilla and Exponential. ``Rand" refers to the random policy, ``Eps" is the $\epsilon$-greedy version of our Algorithm \ref{f:2}, ``C-UCB1" is the cascading bandit algorithm of \cite{z+16}, while ``GL-CDCM" is the one from \cite{liu2018contextual}. Moreover, ``Ind" and ``Dep" are abbreviations for the Independent (Algorithm \ref{f:2}) and Dependent (Algorithm \ref{a:dependent}) algorithms proposed in this paper. Notice that the exponential scenario does not include the baselines ``Eps", ``C-UCB1" and ``GL-CDCM" since those baselines are defined to work only in the vanilla scenario. For each of the two scenarios and each value of $b_t$, we emphasize in bold the best performance.}
\label{tab:msd-ncr-full-mnist0}
\begin{tabular}{c|c|c|c|c|c|c||c|c|c|}
\cline{2-10}
\multicolumn{1}{l|}{}               & \multicolumn{6}{c||}{Vanilla} & \multicolumn{3}{c|}{Exponential} \\ \hline
\multicolumn{1}{|c|}{$b_t$}  & Rand     & Eps         & C-UCB1  & GL-CDCM   & Ind         & Dep     &Rand     & Ind       & Dep                \\ \hline
\multicolumn{1}{|c|}{1}      & 0.00     & 0.95 &       0.84 &    0.98 &      0.97 &       {\bf 1.00}      & 0.00    &  0.97 &       {\bf 1.00}                      \\ \hline
\multicolumn{1}{|c|}{5}      & 0.00     & 0.97 &       0.99 &      0.99 &      0.99 &       {\bf 1.00}      & 0.00    &  0.99 &       {\bf 1.00}                        \\ \hline
\multicolumn{1}{|c|}{10}     & 0.00     & 0.97 &       {\bf 0.99} &    {\bf 0.99} &      {\bf 0.99} &       {\bf 0.99}      & 0.00    &  0.99 &       {\bf 1.00}                        \\ \hline
\multicolumn{1}{|c|}{50}     & 0.00     & {\bf 0.70} &       {\bf 0.70} &   {\bf 0.70} &      {\bf 0.70} &       {\bf 0.70}      & 0.00    &  0.99 &       {\bf 1.00}                        \\ \hline
\multicolumn{1}{|c|}{100}    & {\bf 0.00}     & {\bf 0.00} &       {\bf 0.00} &  {\bf 0.00} &      {\bf 0.00} &       {\bf 0.00}    & 0.00    &  0.99 &       {\bf 1.00}                        \\ \hline
\end{tabular}
\end{table}

\begin{table}[t!]
\smaller
\centering
\caption{Same as in Table \ref{tab:msd-ncr-full-mnist0} for MNIST Dataset with pivot-class $1$.}
\label{tab:msd-ncr-full-mnist1}
\begin{tabular}{c|c|c|c|c|c|c||c|c|c|}
\cline{2-10}
\multicolumn{1}{l|}{}               & \multicolumn{6}{c||}{Vanilla} & \multicolumn{3}{c|}{Exponential} \\ \hline
\multicolumn{1}{|c|}{$b_t$}  & Rand     & Eps         & C-UCB1  & GL-CDCM   & Ind         & Dep     &Rand     & Ind       & Dep                \\ \hline
\multicolumn{1}{|c|}{1}      & 0.00     & 0.91 &       0.93 &    0.99 &      0.98 &       {\bf 1.00}      & 0.00    &  0.98 &       {\bf 1.00}                      \\ \hline
\multicolumn{1}{|c|}{5}      & 0.00     & 0.97 &       0.99 &      0.99 &      0.99 &       {\bf 1.00}      & 0.00    &  0.99 &       {\bf 1.00}                        \\ \hline
\multicolumn{1}{|c|}{10}     & 0.00     & {\bf 0.99} &       {\bf 0.99} &   {\bf 0.99} &      {\bf 0.99} &       {\bf 0.99}      & 0.00    &  0.99 &      {\bf 1.00}                        \\ \hline
\multicolumn{1}{|c|}{50}     & 0.00     & {\bf 0.58} &       {\bf 0.58} &   {\bf 0.58} &      {\bf 0.58} &       {\bf 0.58}       & 0.00    &  0.99 &       {\bf 1.00}                        \\ \hline
\multicolumn{1}{|c|}{100}    & {\bf 0.00}     &  {\bf 0.00} &        {\bf 0.00} &   {\bf 0.00} &       {\bf 0.00} &        {\bf 0.00}    & 0.00    &   0.99 &       {\bf 1.00}                        \\ \hline
\end{tabular}
\end{table}

\begin{table}[t!]
\smaller
\centering
\caption{Same as in Table \ref{tab:msd-ncr-full-mnist0} for MNIST Dataset with pivot-class $2$.}
\label{tab:msd-ncr-full-mist2}
\begin{tabular}{c|c|c|c|c|c|c||c|c|c|}
\cline{2-10}
\multicolumn{1}{l|}{}               & \multicolumn{6}{c||}{Vanilla} & \multicolumn{3}{c|}{Exponential} \\ \hline
\multicolumn{1}{|c|}{$b_t$}  & Rand     & Eps         & C-UCB1  & GL-CDCM   & Ind         & Dep     &Rand     & Ind       & Dep                \\ \hline
\multicolumn{1}{|c|}{1}      & 0.00     & 0.94 &       0.90 &    0.98 &      {\bf 0.99} &       {\bf 0.99}      & 0.00    &  {\bf 0.99} &       {\bf 0.99}                      \\ \hline
\multicolumn{1}{|c|}{5}      & 0.00     & 0.97 &        {\bf 1.00} &       {\bf 1.00} &       {\bf 1.00} &       {\bf 1.00}      & 0.00    &  {\bf 1.00} &       0.99                        \\ \hline
\multicolumn{1}{|c|}{10}     & 0.00     & 0.97 &       {\bf 0.99} &    {\bf 0.99} &      {\bf 0.99} &       {\bf 0.99}      & 0.00    &  {\bf 1.00} &       {\bf 1.00}                        \\ \hline
\multicolumn{1}{|c|}{50}     & 0.00     & {\bf 0.74} &       {\bf 0.74} &   {\bf 0.74} &      {\bf 0.74} &       {\bf 0.74}      & 0.00    &  {\bf 0.99} &       {\bf 0.99}                        \\ \hline
\multicolumn{1}{|c|}{100}    & {\bf 0.00}    &{\bf 0.00} &       {\bf 0.00} &  {\bf 0.00} &      {\bf 0.00} &       {\bf 0.00}    & 0.00    & {\bf 0.99} &      {\bf 0.99}                        \\ \hline
\end{tabular}
\end{table}

\begin{table}[t!]
\smaller
\centering
\caption{Same as in Table \ref{tab:msd-ncr-full-mnist0} for MNIST Dataset with pivot-class $3$.}
\label{tab:msd-ncr-full-mnist3}
\begin{tabular}{c|c|c|c|c|c|c||c|c|c|}
\cline{2-10}
\multicolumn{1}{l|}{}               & \multicolumn{6}{c||}{Vanilla} & \multicolumn{3}{c|}{Exponential} \\ \hline
\multicolumn{1}{|c|}{$b_t$}  & Rand     & Eps         & C-UCB1  & GL-CDCM   & Ind         & Dep     &Rand     & Ind       & Dep                \\ \hline
\multicolumn{1}{|c|}{1}      & 0.00     & 0.93 &       0.87 &     {\bf 0.94} &      {\bf 0.94} &       0.84      & 0.00    &  {\bf 0.94} &       0.85                      \\ \hline
\multicolumn{1}{|c|}{5}      & 0.00     & 0.94 &       0.96 &      {\bf 0.99} &      {\bf 0.99} &       {\bf 0.99}      & 0.00    &  {\bf 0.98} &       0.94                        \\ \hline
\multicolumn{1}{|c|}{10}     & 0.00     & 0.90 &       0.98 &    {\bf 0.99} &      {\bf 0.99} &       {\bf 0.99}      & 0.00    &  {\bf 0.98} &       0.93                        \\ \hline
\multicolumn{1}{|c|}{50}     & 0.00     & -0.66 &       {\bf 0.76} &    {\bf 0.76} &      {\bf 0.76} &       {\bf 0.76}       & 0.00    &  {\bf 0.97} &       0.89                        \\ \hline
\multicolumn{1}{|c|}{100}    & {\bf 0.00}     & -1.00 &       {\bf 0.00} &  {\bf 0.00} &      {\bf 0.00} &       {\bf 0.00}    & 0.00    &  {\bf 0.98} &       0.89                         \\ \hline
\end{tabular}
\end{table}

\begin{table}[t!]
\smaller
\centering
\caption{Same as in Table \ref{tab:msd-ncr-full-mnist0} for MNIST Dataset with pivot-class $4$.}
\label{tab:msd-ncr-full-mnist4}
\begin{tabular}{c|c|c|c|c|c|c||c|c|c|}
\cline{2-10}
\multicolumn{1}{l|}{}               & \multicolumn{6}{c||}{Vanilla} & \multicolumn{3}{c|}{Exponential} \\ \hline
\multicolumn{1}{|c|}{$b_t$}  & Rand     & Eps         & C-UCB1  & GL-CDCM   & Ind         & Dep     &Rand     & Ind       & Dep                \\ \hline
\multicolumn{1}{|c|}{1}      & 0.00     & 0.93 &       0.83 &     {\bf 0.93} &      0.92 &       0.71      & 0.00    &  {\bf 0.92} &       0.71                     \\ \hline
\multicolumn{1}{|c|}{5}      & 0.00     & 0.92 &       {\bf 0.99} &      {\bf 0.99} &      {\bf 0.99} &       0.96      & 0.00    &  {\bf 0.98} &       0.85                        \\ \hline
\multicolumn{1}{|c|}{10}     & 0.00     & 0.92 &       {\bf 0.99} &    {\bf 0.99} &      {\bf 0.99} &       0.98      & 0.00    &  {\bf 0.98} &       0.84                        \\ \hline
\multicolumn{1}{|c|}{50}     & 0.00     & -0.52 &       {\bf 0.75} &    {\bf 0.75} &      {\bf 0.75} &       {\bf 0.75}      & 0.00    &  {\bf 0.97} &       0.78                        \\ \hline
\multicolumn{1}{|c|}{100}    & {\bf 0.00}     & {\bf 0.00} &       {\bf 0.00} &  {\bf 0.00} &      {\bf 0.00} &       {\bf 0.00}    & 0.00    &  {\bf 0.97} &       0.79                        \\ \hline
\end{tabular}
\end{table}

\begin{table}[t!]
\smaller
\centering
\caption{Same as in Table \ref{tab:msd-ncr-full-mnist0} for MNIST Dataset with pivot-class $5$.}
\label{tab:msd-ncr-full-mnist5}
\begin{tabular}{c|c|c|c|c|c|c||c|c|c|}
\cline{2-10}
\multicolumn{1}{l|}{}               & \multicolumn{6}{c||}{Vanilla} & \multicolumn{3}{c|}{Exponential} \\ \hline
\multicolumn{1}{|c|}{$b_t$}  & Rand     & Eps         & C-UCB1  & GL-CDCM   & Ind         & Dep     &Rand     & Ind       & Dep                \\ \hline
\multicolumn{1}{|c|}{1}      & 0.00     & 0.84 &       0.85 &     0.88 &      {\bf 0.89} &       0.68      & 0.00    &  {\bf 0.89} &       0.69                      \\ \hline
\multicolumn{1}{|c|}{5}      & 0.00     & 0.84 &       0.96 &      {\bf 0.98} &      0.97 &       0.96      & 0.00    &  {\bf 0.95} &       0.82                        \\ \hline
\multicolumn{1}{|c|}{10}     & 0.00     & 0.75 &       0.97 &     0.97 &      {\bf 0.98} &       0.97      & 0.00    &  {\bf 0.94} &       0.81                        \\ \hline
\multicolumn{1}{|c|}{50}     & 0.00     & -2.97 &       {\bf 0.82} &    {\bf 0.82} &      {\bf 0.82} &       {\bf 0.82}      & 0.00    &  {\bf 0.93} &       0.73                        \\ \hline
\multicolumn{1}{|c|}{100}    & {\bf 0.00}     & -3.00 &       {\bf 0.00} &  {\bf 0.00} &      {\bf 0.00} &       {\bf 0.00}    & 0.00    &  {\bf 0.92} &       0.72                        \\ \hline
\end{tabular}
\end{table}

\begin{table}[t!]
\smaller
\centering
\caption{Same as in Table \ref{tab:msd-ncr-full-mnist0} for MNIST Dataset with pivot-class $6$.}
\label{tab:msd-ncr-full-mnist6}
\begin{tabular}{c|c|c|c|c|c|c||c|c|c|}
\cline{2-10}
\multicolumn{1}{l|}{}               & \multicolumn{6}{c||}{Vanilla} & \multicolumn{3}{c|}{Exponential} \\ \hline
\multicolumn{1}{|c|}{$b_t$}  & Rand     & Eps         & C-UCB1  & GL-CDCM   & Ind         & Dep     &Rand     & Ind       & Dep                \\ \hline
\multicolumn{1}{|c|}{1}      & 0.00     & 0.94 &       0.97 &     0.97 &      0.95 &       {\bf 1.00}      & 0.00    &  0.95 &       {\bf 1.00}                      \\ \hline
\multicolumn{1}{|c|}{5}      & 0.00     & 0.97 &       {\bf 1.00} &      0.98 &      0.98 &       {\bf 1.00}      & 0.00    &  0.98 &       {\bf 1.00}                        \\ \hline
\multicolumn{1}{|c|}{10}     & 0.00     & 0.96 &       {\bf 0.99} &     {\bf 0.99} &      {\bf 0.99} &       {\bf 0.99}      & 0.00    &  0.99 &       {\bf 1.00}                        \\ \hline
\multicolumn{1}{|c|}{50}     & 0.00     & 0.46 &       {\bf 0.73} &    {\bf 0.73} &      {\bf 0.73} &       {\bf 0.73}      & 0.00    &  0.98 &       {\bf 1.00}                        \\ \hline
\multicolumn{1}{|c|}{100}    & {\bf 0.00}     & {\bf 0.00} &       {\bf 0.00} &  {\bf 0.00} &      {\bf 0.00} &       {\bf 0.00}    & 0.00    &  0.98 &       {\bf 1.00}                        \\ \hline
\end{tabular}
\end{table}

\begin{table}[t!]
\smaller
\centering
\caption{Same as in Table \ref{tab:msd-ncr-full-mnist0} for MNIST Dataset with pivot-class $7$.}
\label{tab:msd-ncr-full-mnist7}
\begin{tabular}{c|c|c|c|c|c|c||c|c|c|}
\cline{2-10}
\multicolumn{1}{l|}{}               & \multicolumn{6}{c||}{Vanilla} & \multicolumn{3}{c|}{Exponential} \\ \hline
\multicolumn{1}{|c|}{$b_t$}  & Rand     & Eps         & C-UCB1  & GL-CDCM   & Ind         & Dep     &Rand     & Ind       & Dep                \\ \hline
\multicolumn{1}{|c|}{1}      & 0.00     & 0.97 &       0.97 &     0.97 &      0.97 &       {\bf 0.98}      & 0.00    &  0.97 &       {\bf 0.99}                      \\ \hline
\multicolumn{1}{|c|}{5}      & 0.00     & 0.97 &       0.99 &      0.99 &      0.99 &       {\bf 1.00}      & 0.00    &  0.99 &       {\bf 1.00}                        \\ \hline
\multicolumn{1}{|c|}{10}     & 0.00     & 0.96 &       {\bf 0.99} &     {\bf 0.99} &      {\bf 0.99} &       {\bf 0.99}      & 0.00    &  0.99 &       {\bf 1.00}                        \\ \hline
\multicolumn{1}{|c|}{50}     & 0.00     & 0.31 &       {\bf 0.65} &     {\bf 0.65} &      {\bf 0.65} &       {\bf 0.65}      & 0.00    &  0.98 &       {\bf 0.99}                        \\ \hline
\multicolumn{1}{|c|}{100}    & {\bf 0.00}     & {\bf 0.00} &       {\bf 0.00} &  {\bf 0.00} &      {\bf 0.00} &       {\bf 0.00}    & 0.00    &  0.98 &       {\bf 0.99}                        \\ \hline
\end{tabular}
\vspace{1.05in}
\end{table}

\begin{table}[]
\smaller
\centering
\caption{Same as in Table \ref{tab:msd-ncr-full-mnist0} for MNIST Dataset with pivot-class $8$.}
\label{tab:msd-ncr-full-mnist8}
\begin{tabular}{c|c|c|c|c|c|c||c|c|c|}
\cline{2-10}
\multicolumn{1}{l|}{}               & \multicolumn{6}{c||}{Vanilla} & \multicolumn{3}{c|}{Exponential} \\ \hline
\multicolumn{1}{|c|}{$b_t$}  & Rand     & Eps         & C-UCB1  & GL-CDCM   & Ind         & Dep     &Rand     & Ind       & Dep                \\ \hline
\multicolumn{1}{|c|}{1}      & 0.00     & 0.89 &       {\bf 0.92} &     0.91 &      0.91 &       0.71      & 0.00    &  {\bf 0.91} &       0.71                     \\ \hline
\multicolumn{1}{|c|}{5}      & 0.00     & 0.83 &       0.94 &      {\bf 0.99} &      0.98 &       0.96      & 0.00    &  {\bf 0.95} &       0.86                        \\ \hline
\multicolumn{1}{|c|}{10}     & 0.00     & 0.80 &       0.97 &     {\bf 0.98} &      {\bf 0.98} &       {\bf 0.98}      & 0.00    &  {\bf 0.95} &       0.84                        \\ \hline
\multicolumn{1}{|c|}{50}     & 0.00     & -1.59 &       {\bf 0.78} &     {\bf 0.78} &      {\bf 0.78} &       {\bf 0.78}      & 0.00    &  {\bf 0.94} &       0.78                        \\ \hline
\multicolumn{1}{|c|}{100}    & {\bf 0.00}     & -2.00 &       {\bf 0.00} &  {\bf 0.00} &      {\bf 0.00} &       {\bf 0.00}    & 0.00    &    {\bf 0.93} &       0.78                        \\ \hline
\end{tabular}
\end{table}

\begin{table}[t!]
\smaller
\centering
\caption{Same as in Table \ref{tab:msd-ncr-full-mnist0} for MNIST Dataset with pivot-class $9$.}
\label{tab:msd-ncr-full-mnist9}
\begin{tabular}{c|c|c|c|c|c|c||c|c|c|}
\cline{2-10}
\multicolumn{1}{l|}{}               & \multicolumn{6}{c||}{Vanilla} & \multicolumn{3}{c|}{Exponential} \\ \hline
\multicolumn{1}{|c|}{$b_t$}  & Rand     & Eps         & C-UCB1  & GL-CDCM   & Ind         & Dep     &Rand     & Ind       & Dep                \\ \hline
\multicolumn{1}{|c|}{1}      & 0.00     & 0.77 &       {\bf 0.80} &     {\bf 0.80} &      0.75 &       0.74      & 0.00    &  {\bf 0.76} &       0.74                      \\ \hline
\multicolumn{1}{|c|}{5}      & 0.00     & 0.66 &       0.97 &      {\bf 0.99} &      {\bf 0.99} &       0.98      & 0.00    &  {\bf 0.92} &       0.89                        \\ \hline
\multicolumn{1}{|c|}{10}     & 0.00     & 0.56 &       0.98 &     {\bf 0.99} &      {\bf 0.99} &       {\bf 0.99}      & 0.00    &   {\bf 0.91} &       0.86                        \\ \hline
\multicolumn{1}{|c|}{50}     & 0.00     & -4.56 &       {\bf 0.72} &     {\bf 0.72} &      {\bf 0.72} &       {\bf 0.72}      & 0.00    &  {\bf 0.88} &       0.81                        \\ \hline
\multicolumn{1}{|c|}{100}    & {\bf 0.00}     & -3.00 &       {\bf 0.00} &  {\bf 0.00} &      {\bf 0.00} &       {\bf 0.00}    & 0.00    &  {\bf 0.87} &       0.81                        \\ \hline
\end{tabular}
\vspace{0.1in}
\end{table}

\begin{figure}[b]
    \centering
    \vspace{0.2in}
    \includegraphics[width=0.45\textwidth]{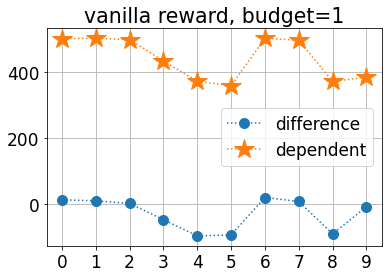} 
    \includegraphics[width=0.45\textwidth]{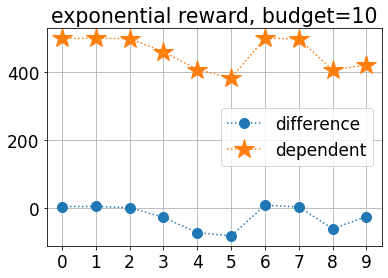} \hspace{-0.10in}
    \caption{MNIST dataset: Correlation across pivot classes $0,\ldots,9$ between the $CR$ performance of Dep (``dependent") and the difference in $CR$ performance between Ind run on the GMM latent space and Ind run without latent space (``difference"). On the left plot is a vanilla scenario with $b_t=1$ or the right an exponential scenario with $b_t=10$. As one can clearly see, on the classes where Dep performs poorly, that is, pivot classes 3, 4, 5, 8, and 9, there is also a substantial degradation in performance for Ind when run on the latent space.
    \vspace{0.2in}
   }
    \label{fig:diff}
\end{figure}

\newpage

\subsection{Further investigations}\label{ssa:further}

In order to further understand the poor performance of Dep in the MNIST classes $3$, $4$, $5$, $8$, and $9$ (Tables \ref{tab:msd-ncr-full-mnist3}-- \ref{tab:msd-ncr-full-mnist5}, \ref{tab:msd-ncr-full-mnist8}, and \ref{tab:msd-ncr-full-mnist9}, respectively), we conducted a small investigation to see to what extent the latent space representation can be deemed responsible for this performance.

We run on the MNIST dataset the independent algorithm Ind on the same GMM-generated latent space on which we ran Dep, and optimized the number $d'$ of centroids as in the tuning of Dep. We then compared the results to Dep as reported in Tables \ref{tab:msd-ncr-full-mnist0}--\ref{tab:msd-ncr-full-mnist9}.

Figure \ref{fig:diff} collects the outcome of this comparison on two relevant scenarios, vanilla with $b_t=1$ and exponential with $b_t=10$. In the x-axis of the two plots are the pivot classes $0,\ldots,9$, on the y-axis are the final $CR$ performances. As one can clearly see from both plots, when Dep performs poorly (classes 3, 4, 5, 8, and 9), it is also the case that the difference in performance between Ind with GMM and Ind without GMM becomes negative, that is to say, those pivot classes for MNIST are the {\em same} classes on which one can observe performance degradation when the GMM-latent space representation is added to Ind.

Though a more thorough investigation on the role of the latent space has to be performed, this finding by itself gives a strong support to the claim that it is indeed the GMM-based latent space that hinders the performance of the bandit algorithms in some cases.

\end{document}